\documentclass[10pt,twocolumn,letterpaper]{article}

\usepackage{cvpr}
\usepackage{times}
\usepackage{epsfig}
\usepackage{graphicx}
\usepackage{amsmath}
\usepackage{amssymb}
\usepackage[noblocks]{authblk}
\usepackage{cite}

\usepackage[pagebackref=true,breaklinks=true,letterpaper=true,colorlinks,bookmarks=false]{hyperref}

 \cvprfinalcopy 

\newcommand{\first}[1] {\textcolor[rgb]{1.0,0.0,0.0}{{#1}}}
\newcommand{\second}[1] {\textcolor[rgb]{0.0,0.0,1.0}{{#1}}}
\newcommand{\third}[1] {\textcolor[rgb]{0.0,0.6,0.6}{{#1}}}

\def\cvprPaperID{2140} 

\ifcvprfinal\fi
\begin{document}

\title{Progressive Image Deraining Networks: A Better and Simpler Baseline}

\author[1]{Dongwei Ren}
\author[2]{Wangmeng Zuo}
\author[1]{Qinghua Hu}
\author[1]{Pengfei Zhu}
\author[3]{Deyu Meng}

\affil[1]{College of Intelligence and Computing, Tianjin University, Tianjin, China}
\affil[2]{School of Computer Science and Technology, Harbin Institute of Technology, Harbin, China }
\affil[3]{Xi'an Jiaotong University, Xi'an, China}

\maketitle

\begin{abstract}
%
Along with the deraining performance improvement of deep networks, their structures and learning become more and more complicated and diverse, making it difficult to analyze the contribution of various network modules when developing new deraining networks.
To handle this issue, this paper provides a better and simpler baseline deraining network by considering network architecture, input and output, and loss functions.
Specifically, by repeatedly unfolding a shallow ResNet, progressive ResNet (PRN) is proposed to take advantage of recursive computation.
A recurrent layer is further introduced to exploit the dependencies of deep features across stages, forming our progressive recurrent network (PReNet).
Furthermore, intra-stage recursive computation of ResNet can be adopted in PRN and PReNet to notably reduce network parameters with unsubstantial degradation in deraining performance.
For network input and output, we take both stage-wise result and original rainy image as input to each ResNet and finally output the prediction of {residual image}.
As for loss functions, single MSE or negative SSIM losses are sufficient to train PRN and PReNet.
Experiments show that PRN and PReNet perform favorably on both synthetic and real rainy images.
Considering its simplicity, efficiency and effectiveness, our models are expected to serve as a suitable baseline in future deraining research.
The source codes are available at \url{https://github.com/csdwren/PReNet}.
%
%
%
%
%
\end{abstract}

\section{Introduction}
Rain is a common weather condition, and has severe adverse effect on not only human visual perception but also the performance of various high level vision tasks such as image classification, object detection, and video surveillance~\cite{kang2012automatic,fu2018lightweight}.
Single image deraining aims at restoring clean background image from a rainy image, and has drawn considerable recent research attention.
For example, several traditional optimization based methods \cite{chen2013generalized,li2016rain,luo2015removing,gu2017joint} have been suggested for modeling and separating rain streaks from background clean image.
However, due to the complex composition of rain and background layers, image deraining remains a challenging ill-posed problem.

\begin{figure}[!tb]\footnotesize
\centering
\setlength{\tabcolsep}{1pt}
\begin{tabular}{ccccccc}
\!\!\!\!\!
 \includegraphics[width=.24\textwidth]{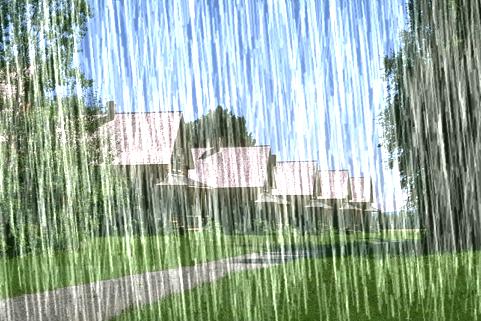} &
 \includegraphics[width=.24\textwidth]{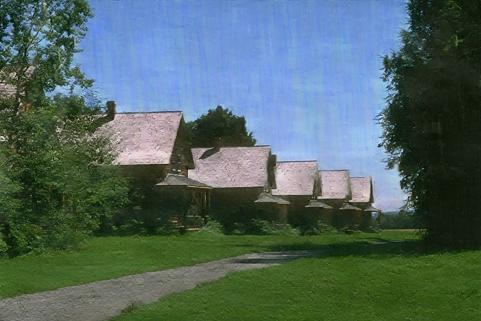} \\
 Rainy image & RESCAN~\cite{li2018recurrent} \\
 \!\!\!\!\!
 \includegraphics[width=.24\textwidth]{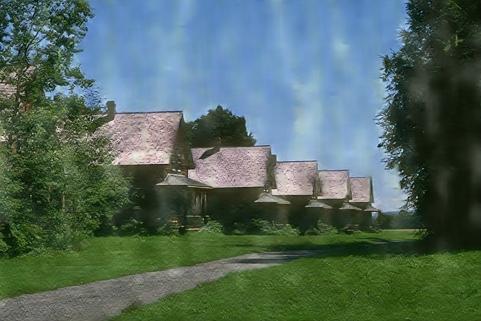} &
 \includegraphics[width=.24\textwidth]{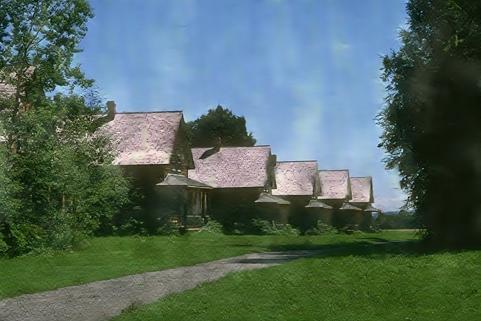}\\
  $t = 1$ & $t = 2$\\
  \!\!\!\!\!
 \includegraphics[width=.24\textwidth]{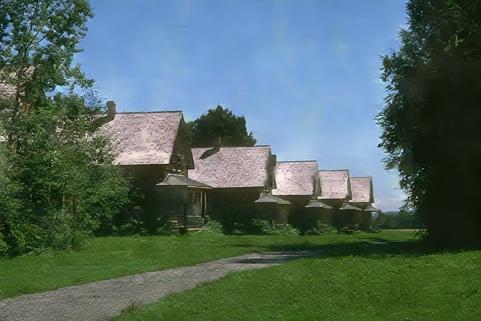} &
 \includegraphics[width=.24\textwidth]{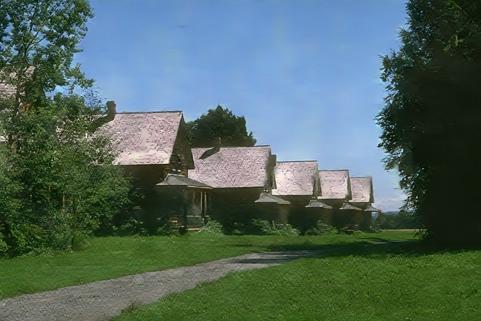} \\
  $t = 4$ & $t = 6$  \\
\end{tabular}
   \caption{Deraining results by RESCAN \cite{li2018recurrent} and PReNet ($T=6$) at stage $t = 1,2,4,6$, respectively. }
\label{fig:states results}
\end{figure}
\begin{figure*}[!ht]\footnotesize
	\centering
	\begin{tabular}{lcccccc}
		\ \ \
		\includegraphics[width=.1\textwidth]{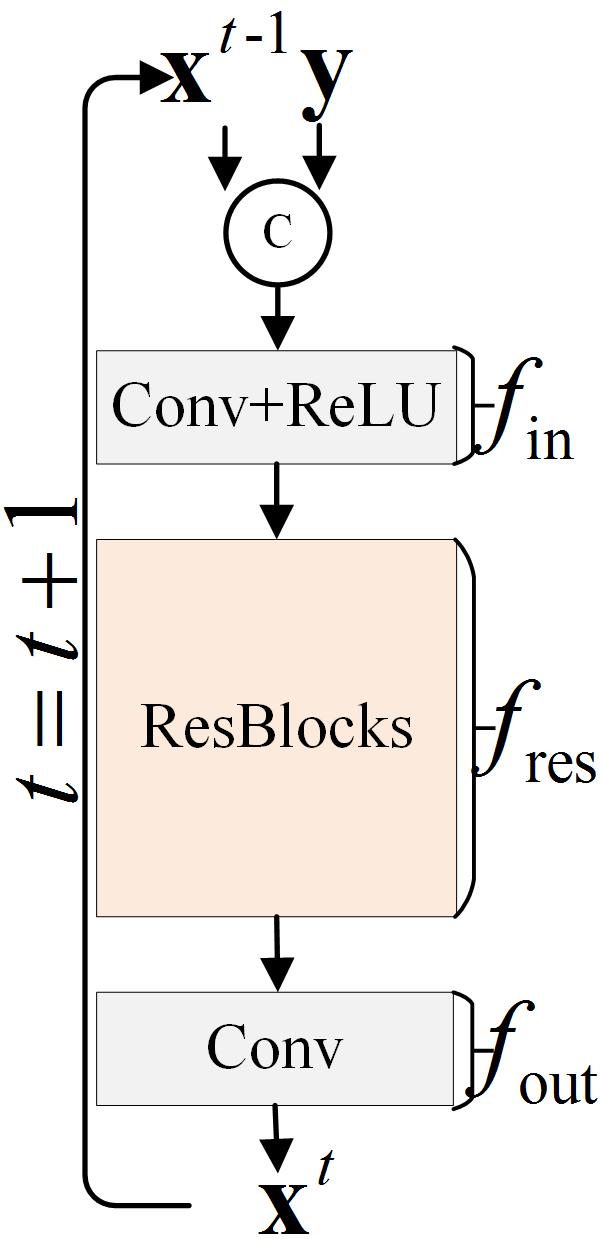} \ \ \ &
		\includegraphics[width=.8\textwidth]{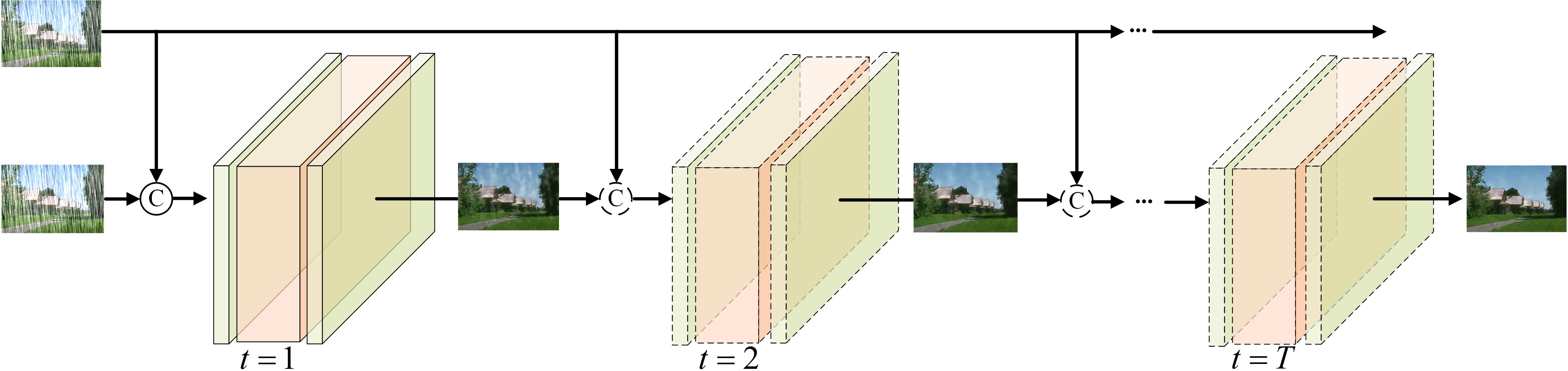}\\
		\multicolumn{2}{c}{(a) PRN and the illustration of PRN with $T$ stages recursion}\\
		\ \ \
		\includegraphics[width=.1\textwidth]{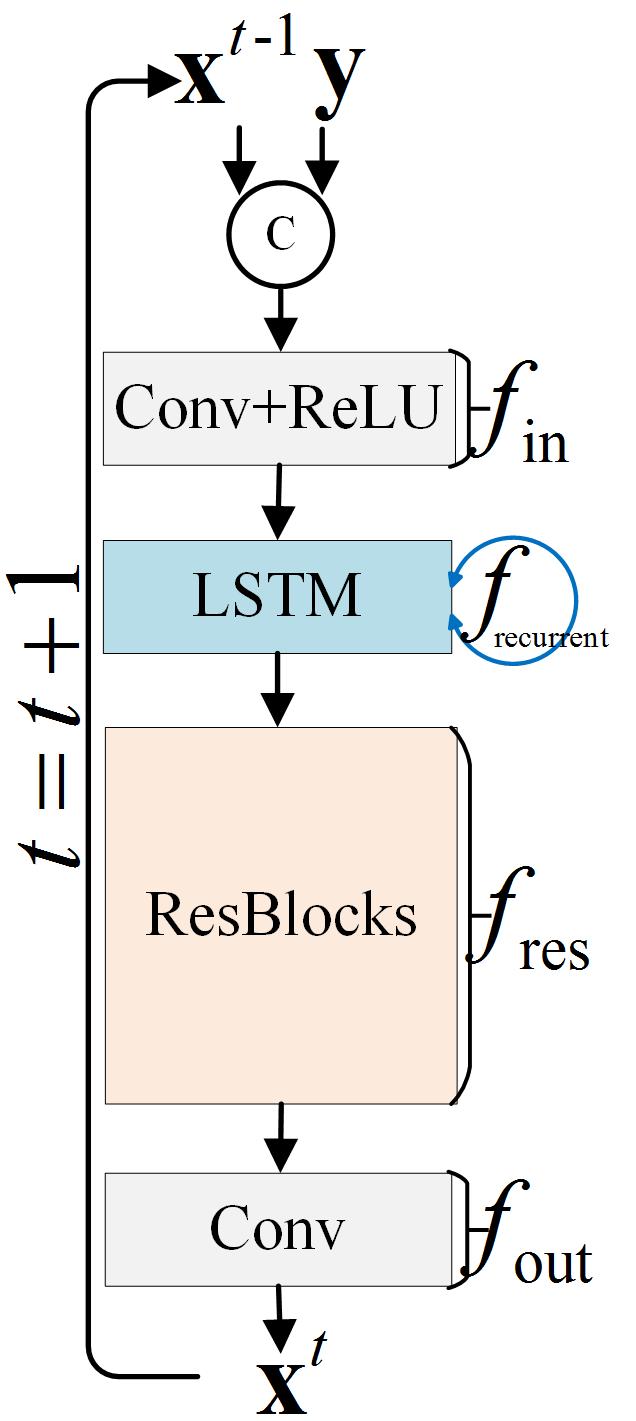} \ \ \ &
		\includegraphics[width=.8\textwidth]{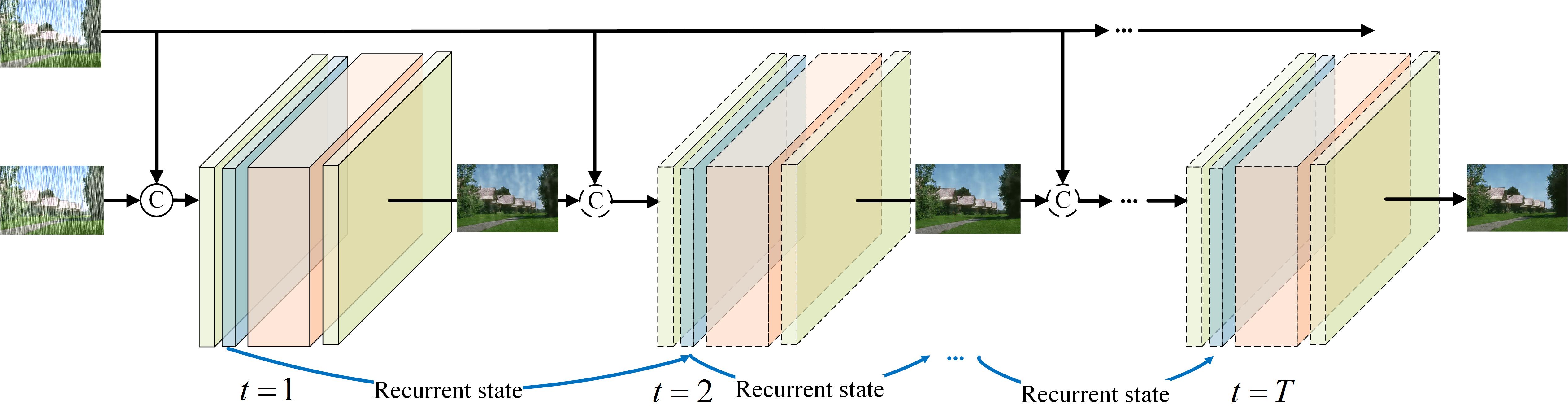}\\
		\multicolumn{2}{c}{(b) PReNet and the illustration of PReNet with $T$ stages recursion}\\
	\end{tabular}
	\caption{The architectures of progressive networks, where $f_{\text{in}}$ is a convolution layer with \emph{ReLU}, $f_{\text{res}}$ is \emph{ResBlocks}, $f_{\text{out}}$ is a convolution layer, $f_{\text{recurrent}}$ is a convolutional LSTM and $\copyright$ is a \emph{concat} layer. $f_{\text{res}}$ can be implemented as conventional ResBlocks or recursive ResBlocks as shown in Fig. \ref{fig:recursive resnet}. }
	\label{fig:network}
\end{figure*}

Driven by the unprecedented success of deep learning in low level vision \cite{zhang2017beyond,dong2016image,ledig2017photo,kim2016deeply,tai2017image}, recent years have also witnessed the rapid progress of deep convolutional neural network (CNN) in image deraining.
In~\cite{fu2017clearing}, Fu \etal show that it is difficult to train a CNN to directly predict background image from rainy image, and utilize a 3-layer CNN to remove rain streaks from high-pass detail layer instead of the input image.
Subsequently, other formulations are also introduced, such as residual learning for predicting rain steak layer~\cite{li2018recurrent}, joint detection and removal of rain streaks~\cite{yang2017deep}, and joint rain density estimation and deraining~\cite{zhang2018density}.

On the other hand, many modules are suggested to constitute different deraining networks, including residual blocks \cite{fu2017removing,he2016deep}, dilated convolution~\cite{yu2016multi,yang2017deep}, dense blocks~\cite{zhang2018density}, squeeze-and-excitation~\cite{li2018recurrent}, and recurrent layers~\cite{li2018recurrent,qian2018attentive}.
Multi-stream~\cite{zhang2018density} and multi-stage~\cite{li2018recurrent} networks are also deployed to capture multi-scale characteristics and to remove heavy rain.
Moreover, several models are designed to improve computational efficiency by utilizing lightweight networks in a cascaded scheme~\cite{fan2018residual} or a Laplacian pyramid framework \cite{fu2018lightweight}, but at the cost of obvious degradation in deraining performance.
To sum up, albeit the progress of deraining performance, the structures of deep networks become more
and more complicated and diverse.
As a result, it is difficult to analyze the contribution of various modules and their combinations, and to develop new models by introducing modules to existing deeper and complex deraining networks.

In this paper, we aim to present a new baseline network for single image deraining to demonstrate that:
(i) by combining only a few modules, a better and simpler baseline network can be constructed and achieve noteworthy performance gains over state-of-the-art deeper and complex deraining networks,
(ii) unlike~\cite{fu2017clearing}, the improvement of deraining networks may ease the difficulty of training CNNs to directly recover clean image from rainy image.
Moreover, the simplicity of baseline network makes it easier to develop new deraining models by introducing other network modules or modifying the existing ones.

To this end, we consider the network architecture, input and output, and loss functions to form a better and simpler baseline networks.
In terms of network architecture, we begin with a basic shallow residual network (ResNet) with five residual blocks (ResBlocks).
Then, progressive ResNet (PRN) is introduced by recursively unfolding the ResNet into multiple stages without the increase of model parameters (see Fig.~\ref{fig:network}(a)).
Moreover, a recurrent layer \cite{hochreiter1997long,xingjian2015convolutional} is introduced to exploit the dependencies of deep features across recursive stages to form the PReNet in Fig.~\ref{fig:network}(b).
From Fig.~\ref{fig:states results}, a 6-stage PReNet can remove most rain streaks at the first stage, and then remaining rain streaks can be progressively removed, leading to promising deraining quality at the final stage.
%
%
Furthermore, PRN$_r$ and PReNet$_r$ are presented by adopting intra-stage recursive unfolding of only one ResBlock, which reduces network parameters only at the cost of unsubstantial performance degradation.

Using PRN and PReNet, we further investigate the effect of network input/output and loss function.
In terms of network input, we take both stage-wise result and original rainy image as input to each ResNet, and empirically find that the introduction of original image does benefit deraining performance.
In terms of network output, we adopt the residual learning formulation by predicting rain streak layer, and find that it is also feasible to directly learn a PRN or PReNet model for predicting clean background from rainy image.
Finally, instead of hybrid losses with careful hyper-parameters tuning~\cite{fan2018residual,fu2017removing}, a single negative SSIM \cite{wang2004image} or MSE loss  can readily train PRN and PReNet with favorable deraining performance.

Comprehensive experiments have been conducted to evaluate our baseline networks PRN and PReNet.
On four synthetic datasets, our PReNet and PRN are computationally very efficient, and achieve much better quantitative and qualitative deraining results in comparison with the state-of-the-art methods.
In particular, on heavy rainy dataset Rain100H \cite{yang2017deep}, the performance gains by our PRN and PReNet are still significant.
The visually pleasing deraining results on real rainy images and videos have also validated the generalization ability of the trained PReNet and PRN models.

The contribution of this work is four-fold:
\begin{itemize}
\vspace{-0.06in}
\item Baseline deraining networks, \ie, PRN and PReNet, are proposed, by which better and simpler networks can work well in removing rain streaks, and provide a suitable basis to future studies on image deraining.
\vspace{-0.06in}
\item By taking advantage of intra-stage recursive computation, PRN$_r$ and PReNet$_r$ are also suggested to reduce network parameters while maintaining state-of-the-art deraining performance.
\vspace{-0.06in}
\item Using PRN and PReNet, the deraining performance can be further improved by taking both stage-wise result and original rainy image as input to each ResNet, and our progressive networks can be readily trained with single negative SSIM or MSE loss.
\vspace{-0.06in}
\item Extensive experiments show that our baseline networks are computationally very efficient, and perform favorably against state-of-the-arts on both synthetic and real rainy images.
\end{itemize}

%

\section{Related Work}\label{sec:related_work}
In this section, we present a brief review on two representative types of deraining methods, \ie, traditional optimization-based and deep network-based ones.

\subsection{Optimization-based Deraining Methods}
In general, a rainy image can be formed as the composition of a clean background image layer and a rain layer.
On the one hand, linear summation is usually adopted as the composition model~\cite{chen2013generalized,li2016rain,zhu2017joint}.
Then, image deraining can be formulated by incorporating with proper regularizers on both background image and rain layer, and solved by specific optimization algorithms.
Among these methods, Gaussian mixture model (GMM)~\cite{li2016rain}, sparse representation~\cite{zhu2017joint}, and low rank representation~\cite{chen2013generalized} have been adopted for modeling background image or a rain layers.
%
%
%
Based on linear summation model, optimization-based methods have been also extended for video deraining \cite{garg2004detection,jiang2017novel,jiang2018fastderain,kim2015video,li2018video}.
On the other hand, screen blend model \cite{luo2015removing,ren2018simultaneous} is assumed to be more realistic for the composition of rainy image, based on which Luo \etal \cite{luo2015removing} use
the discriminative dictionary learning to separate rain streaks by enforcing the two layers share fewest dictionary atoms.
However, the real composition generally is more complicated and the regularizers are still insufficient in characterizing background and rain layers, making optimization-based methods remain limited in deraining performance.


\subsection{Deep Network-based Deraining Methods}
When applied deep network to single image deraining, one natural solution is to learn a direct mapping to predict clean background image $\mathbf{x}$ from rainy image $\mathbf{y}$.
However, it is suggested that plain fully convolutional networks (FCN) are ineffective in learning the direct mapping \cite{fu2017clearing,fu2017removing}.
Instead, Fu \etal \cite{fu2017clearing,fu2017removing} apply a low-pass filter to decompose $\mathbf{y}$ into a base layer $\mathbf{y}_{\text{base}}$ and a detail layer $\mathbf{y}_{\text{detail}}$.
By assuming $\mathbf{y}_{\text{base}} \approx \mathbf{x}_{\text{base}}$, FCNs are then deployed to predict $\mathbf{x}_{\text{detail}}$ from $\mathbf{y}_{\text{detail}}$.
In contrast, Li \etal \cite{li2018recurrent} adopt the residual learning formulation to predict rain layer $\mathbf{y} - \mathbf{x}$ from $\mathbf{y}$.
More complicated learning formulations, such as joint detection and removal of rain streaks~\cite{yang2017deep}, and joint rain density estimation and deraining~\cite{zhang2018density}, are also suggested.
And adversarial loss is also introduced to enhance the texture details of deraining result \cite{zhang2017image,qian2018attentive}.
In this work, we show that the improvement of deraining networks actually eases the difficulty of learning, and it is also feasible to train PRN and PReNet to learn either direct or residual mapping.

For the architecture of deraining network, Fu \etal first adopt a shallow CNN \cite{fu2017clearing} and then a deeper ResNet \cite{fu2017removing}.
In \cite{yang2017deep}, a multi-task CNN architecture is designed for joint detection and removal of rain streaks, in which contextualized dilated convolution and recurrent structure are adopted to handle multi-scale and heavy rain steaks.
Subsequently, Zhang \etal \cite{zhang2018density} propose a density aware multi-stream densely connected CNN for joint estimating rain density and removing rain streaks.
In~\cite{qian2018attentive}, attentive-recurrent network is developed for single image raindrop removal.
Most recently, Li \etal \cite{li2018recurrent} recurrently utilize dilated CNN and squeeze-and-excitation blocks to remove heavy rain streaks.
In comparison to these deeper and complex networks, our work incorporates ResNet, recurrent layer and multi-stage recursion to constitute a better, simpler and more efficient deraining network.

Besides, several lightweight networks, \eg, cascaded scheme \cite{fan2018residual} and Laplacian pyrimid framework \cite{fu2018lightweight} are also developed to improve computational efficiency but at the cost of obvious performance degradation.
As for PRN and PReNet, we further introduce intra-stage recursive computation to reduce network parameters while maintaining state-of-the-art deraining performance, resulting in our PRN$_r$ and PReNet$_r$ models.

\section{Progressive Image Deraining Networks}
In this section, progressive image deraining networks are presented by considering network architecture, input and output, and loss functions.
To this end, we first describe the general framework of progressive networks as well as input/output, then implement the network modules, and finally discuss the learning objectives of progressive networks.
%

\subsection{Progressive Networks}
A simple deep network generally cannot succeed in removing rain streaks from rainy images \cite{fu2017clearing,fu2017removing}.
Instead of designing deeper and complex networks, we suggest to tackle the deraining problem in multiple stages, where a shallow ResNet is deployed at each stage.
One natural multi-stage solution is to stack several sub-networks, which inevitably leads to the increase of network parameters and susceptibility to over-fitting.
In comparison, we take advantage of inter-stage recursive computation \cite{kim2016deeply,tai2017image,li2018recurrent} by requiring multiple stages share the same network parameters.
Besides, we can incorporate intra-stage recursive unfolding of only 1 ResBlock to significantly reduce network parameters with graceful degradation in deraining performance.


%
%

\vspace{-.1in}
\subsubsection{Progressive Residual Network}
%
%

We first present a progressive residual network (PRN) as shown in Fig.~\ref{fig:network}(a).
In particular, we adopt a basic ResNet with three parts: (i) a convolution layer $f_{\text{in}}$ receives network inputs, (ii) several residual blocks (\emph{ResBlocks}) $f_{\text{res}}$ extract deep representation, and (iii) a convolution layer $f_{\text{out}}$ outputs deraining results.
The inference of PRN at stage $t$ can be formulated as
\begin{equation} \label{eq:prn}
\begin{aligned}
&\mathbf{x}^{t-0.5}     = f_{\text{in}}(\mathbf{x}^{t-1}, \mathbf{y}), \\
&\mathbf{x}^{t}       = f_{\text{out}}(f_{\text{res}}(\mathbf{x}^{t-0.5})),\\
\end{aligned}
\end{equation}
where $f_{\text{in}}$, $f_{\text{res}}$ and $f_{\text{out}}$ are stage-invariant, \ie, network parameters are reused across different stages.

We note that $f_{\text{in}}$ takes the concatenation of the current estimation $\mathbf{x}^{t-1}$ and rainy image $\mathbf{y}$ as input.
In comparison to only $\mathbf{x}^{t-1}$ in \cite{li2018recurrent}, the inclusion of $\mathbf{y}$ can further improve the deraining performance.
The network output can be the prediction of either rain layer or clean background image.
Our empirical study show that, although predicting rain layer performs moderately better, it is also possible to learn progressive networks for predicting background image.
%

%


\vspace{-.1in}
\subsubsection{Progressive Recurrent Network}

We further introduce a recurrent layer into PRN, by which feature dependencies across stages can be propagated to facilitate rain streak removal, resulting in our progressive recurrent network (PReNet).
The only difference between PReNet and PRN is the inclusion of recurrent state $\mathbf{s}^{t}$,
\begin{equation} \label{eq:prenet}
\begin{aligned}
&\mathbf{x}^{t-0.5}     = f_{\text{in}}(\mathbf{x}^{t-1}, \mathbf{y}), \\
&\mathbf{s}^{t}       = f_{\text{recurrent}} (\mathbf{s}^{t-1}, \mathbf{x}^{t-0.5}),\\
&\mathbf{x}^{t}       = f_{\text{out}}(f_{\text{res}}(\mathbf{s}^{t})),\\
\end{aligned}
\end{equation}
where the recurrent layer $f_{\text{recurrent}}$ takes both $\mathbf{x}^{t-0.5}$ and the recurrent state $\mathbf{s}^{t-1}$ as input at stage $t-1$.
$f_{\text{recurrent}}$ can be implemented using either convolutional Long Short-Term Memory (LSTM)~\cite{hochreiter1997long,xingjian2015convolutional} or convolutional Gated Recurrent Unit (GRU)~\cite{cho2014learning}.
In PReNet, we choose LSTM due to its empirical superiority in image deraining.

The architecture of PReNet is shown in Fig.~\ref{fig:network}(b).
By unfolding PReNet with $T$ recursive stages, the deep representation that facilitates rain streak removal are propagated by recurrent states.
The deraining results at intermediate stages in Fig.~\ref{fig:states results} show that the heavy rain streak accumulation can be gradually removed stage-by-stage.

\subsection{Network Architectures}
\begin{figure}[!ht]\footnotesize
	\centering
	\begin{tabular}{ccccccc}
		\!\!\!\!\!\!
		\includegraphics[width=.21\textwidth]{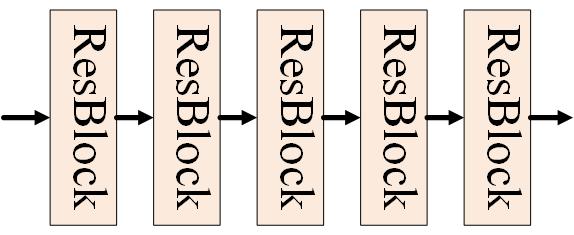}   &
		
		\includegraphics[width=.21\textwidth]{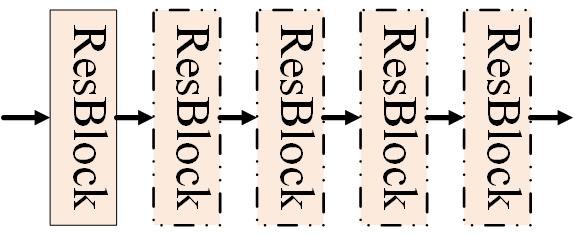}\\
		(a) Conventional ResBlocks  & (b) Recursive ResBlocks\\
		
	\end{tabular}
	\caption{Implementations of $f_{\text{res}}$: (a) conventinal ResBlocks and (b) recursive ResBlocks where one ResBlock is recursively unfolded five times.  }
	\label{fig:recursive resnet}
\end{figure}

We hereby present the network architectures of PRN and PReNet.
All the filters are with size $3\times 3$ and padding  $1\times 1$.
Generally, $f_{\text{in}}$ is a 1-layer convolution with ReLU nonlinearity \cite{nair2010rectified}, $f_{\text{res}}$ includes 5 {ResBlocks} and $f_{\text{out}}$ is also a 1-layer convolution.
Due to the concatenation of 3-channel RGB $\mathbf{y}$ and 3-channel RGB $\mathbf{x}^{t-1}$, the convolution in $f_{\text{in}}$ has 6 and 32 channels for input and output, respectively.
$f_{\text{out}}$ takes the output of $f_{\text{res}}$ (or $f_{\text{recurrent}}$) with 32 channels as input and outputs 3-channel RGB image for PRN (or PReNet).
In $f_\text{recurrent}$,  all the convolutions in LSTM have 32 input channels and 32 output channels.
$f_{\text{res}}$ is the key component to extract deep representation for rain streak removal, and we provide two implementations, \ie, conventional ResBlocks shown in Fig.~\ref{fig:recursive resnet}(a) and recursive ResBlocks shown in Fig.~\ref{fig:recursive resnet}(b).

\noindent \textbf{Conventional ResBlocks:}
As shown in Fig.~\ref{fig:recursive resnet}(a), we first implement $f_{\text{res}}$ with 5 ResBlocks as its conventional form, \ie, each ResBlock includes 2 convolution layers followed by ReLU \cite{nair2010rectified}.
All the convolution layers receive 32-channel features without downsampling or upsamping operations.
Conventional ResBlocks are adopted in PRN and PReNet.

\vspace{0.03in}
\noindent \textbf{Recursive ResBlocks:}
Motivated by \cite{kim2016deeply,tai2017image}, we also implement $f_{\text{res}}$ by recursively unfolding one ResBlock 5 times, as shown in Fig.~\ref{fig:recursive resnet}(b).
Since network parameters mainly come from ResBlocks, the intra-stage recursive computation leads to a much smaller model size, resulting in PRN$_r$ and PReNet$_r$.
We have evaluated the performance of recursive ResBlocks in Sec.~\ref{sec:synthetic}, indicating its elegant tradeoff between model size and deraining performance.

\subsection{Learning Objective}
Recently, hybrid loss functions, \eg, MSE+SSIM \cite{fan2018residual}, $\ell_1$+SSIM \cite{fu2018lightweight} and even adversarial loss \cite{zhang2017image}, have been widely adopted for training deraining networks, but incredibly increase the burden of hyper-parameter tuning.
In contrast, benefited from the progressive network architecture, we empirically find that a single loss function, \eg, MSE loss or negative SSIM loss \cite{wang2004image}, is sufficient to train PRN and PReNet.

For a model with $T$ stages, we have $T$ outputs, \ie, $\mathbf{x}^1$, $\mathbf{x}^2$,..., $\mathbf{x}^T$.
By only imposing supervision on the final output $\mathbf{x}^T$, the MSE loss is
\begin{equation}\label{eq:mse loss}
\mathcal{L} =  \|\mathbf{x}^T - \mathbf{x}^{gt}\|^2,
\end{equation}
and the negative SSIM loss is
\begin{equation}\label{eq:ssim loss}
\mathcal{L} = - \text{SSIM}\left(\mathbf{x}^T , \mathbf{x}^{gt}\right),
\end{equation}
where $\mathbf{x}^{gt}$ is the corresponding ground-truth clean image.
It is worth noting that, our empirical study shows that negative SSIM loss outperforms MSE loss in terms of both PSNR and SSIM.

Moreover, recursive supervision can be imposed on each intermediate result,
\begin{equation}\label{eq:multi loss}
\mathcal{L} = - \sum\nolimits_{t=1}^{T} \lambda_t \text{SSIM}\left(\mathbf{x}^{t} , \mathbf{x}^{gt}\right),
\end{equation}
where $\lambda_t$ is the tradeoff parameter for stage $t$.
Experimental result in Sec.~\ref{sec:training loss} shows that recursive supervision cannot achieve any performance gain when $t = T$, but can be adopted to generate visually satisfying result at early stages.


%
%

\section{Experimental Results}

In this section, we first conduct ablation studies to verify the main components of our methods, then quantitatively and qualitatively evaluate progressive networks, and finally assess PReNet on real rainy images and video.
All the source code, pre-trained models and results can be found at \url{https://github.com/csdwren/PReNet}.

Our progressive networks are implemented using Pytorch \cite{paszke2017automatic}, and are trained on a PC equipped with two NVIDIA GTX 1080Ti GPUs.
In our experiments, all the progressive networks share the same training setting.
The patch size is $100 \times 100$, and the batch size is 18.
The ADAM \cite{kingma2014adam} algorithm is adopted to train the models with an initial learning rate $1\times 10^{-3}$, and ends after 100 epochs.
When reaching 30, 50 and 80 epochs, the learning rate is decayed by multiplying $0.2$.

\subsection{Ablation Studies}
All the ablation studies are conducted on a heavy rainy dataset \cite{yang2017deep} with 1,800 rainy images for training and 100 rainy images (Rain100H) for testing.
However, we find that 546 rainy images from the 1,800 training samples have the same background contents with testing images.
Therefore, we exclude these 546 images from training set, and train all our models on the remaining 1,254 training images.

\subsubsection{Loss Functions}\label{sec:training loss}
Using PReNet ($T=6$) as an example, we discuss the effect of loss functions on deraining performance, including MSE loss, negative SSIM loss, and recursive supervision loss.

\begin{figure*}[!htb]\footnotesize
	\centering
	\setlength{\tabcolsep}{1pt}
	
	\begin{tabular}{cccccccccccccc}
		\includegraphics[width=.24\textwidth]{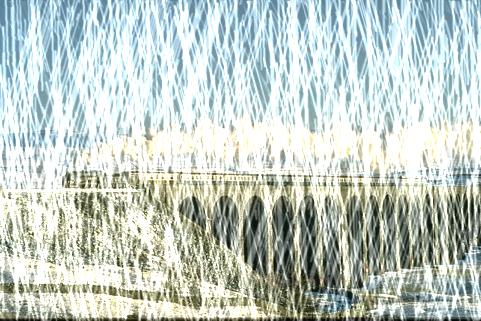}  &
		\includegraphics[width=.24\textwidth]{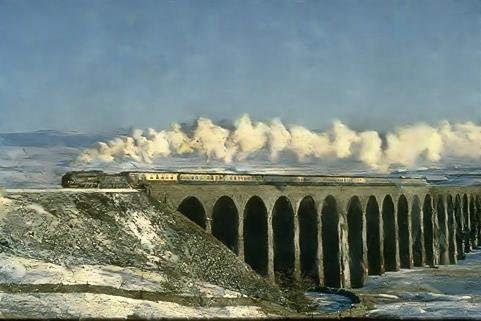} &
		\includegraphics[width=.24\textwidth]{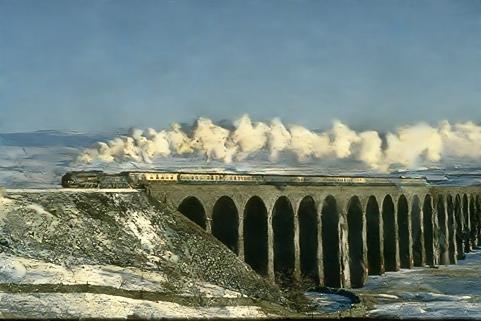}&
		\includegraphics[width=.24\textwidth]{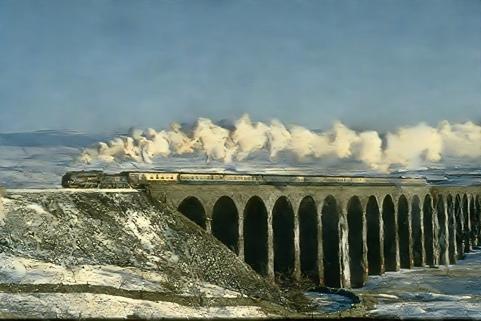}\\
		(a) Rainy image & (b) PReNet-MSE & (c) PReNet-SSIM & (d) PReNet-RecSSIM \\

	\end{tabular}
	\caption{Visual quality comparison of PReNet models trained by different loss functions, including single MSE loss (PReNet-MSE), single negative SSIM loss (PReNet-SSIM) and recursive negative SSIM supervision (PReNet-RecSSIM). }
	\label{fig:results loss}
\end{figure*}

\vspace{.05in}
\noindent \textbf{Negative~SSIM~v.s.~MSE.}
We train two PReNet models by minimizing MSE loss (PReNet-MSE) and negative SSIM loss (PReNet-SSIM), and Table~\ref{table:ablation loss} lists their PSNR and SSIM values on Rain100H.
Unsurprisingly, PReNet-SSIM outperforms PReNet-MSE in terms of SSIM.
We also note that PReNet-SSIM even achieves higher PSNR, partially attributing to that PReNet-MSE may be inclined to get trapped into poor solution.
As shown in Fig. \ref{fig:results loss}, the deraining result by PReNet-SSIM is also visually more plausible than that by PReNet-MSE.
Therefore, negative SSIM loss is adopted as the default in the following experiments.

\begin{table}[!htb]\small
	\caption{Comparison of PReNet ($T$ = 6) with different loss functions. }
	\centering
	\begin{tabular}{c|cc|cccccc}
		\hline
		
		\hline
		Loss    & PReNet-MSE  &  PReNet-SSIM  & PReNet-RecSSIM \\
		\hline
		PSNR    & 29.08       &  29.32        & 29.12          \\
		SSIM    & 0.880       &  0.898        & 0.895          \\
		\hline
		
		\hline
	\end{tabular}
	\label{table:ablation loss}
\end{table}

\begin{table*}[!htb]\small
	\caption{Comparison of PReNet models with different $T$ stages.  }
	\setlength{\tabcolsep}{4pt}
	\centering
	\begin{tabular}{c|cccccccc}
		\hline
		
		\hline
		Model & $\text{PReNet}_2$ & $\text{PReNet}_3$ & $\text{PReNet}_4$ & $\text{PReNet}_5$ & $\text{PReNet}_6$ & $\text{PReNet}_7$ \\
		\hline
		PSNR       & 27.27   &  28.01    &  28.60     &  28.92 & 29.32 &29.24            \\
		SSIM       & 0.882   & 0.885     &  0.890     &  0.895  &0.898 &0.898            \\
		\hline
		
		\hline
	\end{tabular}
	\label{table:ablation stages}
\end{table*}

\begin{table}[!htb]\small
	\caption{Comparisons of PReNet variants for ablation studies. 
PReNet$_x$, PReNet-LSTM, and PReNet-GRU learn direct mapping for predicting background image. 
In particular, PReNet$_x$ only takes current deraining result $\mathbf{x}^{t-1}$ as input, the recurrent layers in PReNet-LSTM and PReNet-GRU are implemented using LSTM and GRU, respectively.
PReNet is the final model by adopting residual learning and LSTM recurrent layer, and taking $\bf y$ and $\mathbf{x}^{t-1}$ as input. }
	\centering
	\setlength{\tabcolsep}{4pt}
	\begin{tabular}{c|c|cc|ccccc}
		\hline
		
		\hline
		Model      &  PReNet$_x$    &  PReNet-LSTM & PReNet-GRU  & PReNet  \\
		\hline
		PSNR       & 28.91    & 29.32        &  29.08         &  29.46         \\
		SSIM       & 0.895    & 0.898        &  0.896        &  0.899        \\
		\hline
		
		\hline
	\end{tabular}
	\label{table:ablation more}
\end{table}

\begin{table}[!htb]\small
	\caption{Effect of recursive ResBlocks. PRN and PReNet contain 5 ResBlocks. PRN$_r$ and PReNet$_r$ unfold 1 ResBlock 5 times. }
	\centering
	\setlength{\tabcolsep}{8pt}
	\begin{tabular}{c|cc|cccccc}
		\hline
		
		\hline
		Model          &  PRN & PReNet &  PRN$_r$ & PReNet$_r$  \\
		\hline
		PSNR           & 28.07&  29.46 &  27.43   & 28.98     \\
		SSIM           & 0.884&  0.899 &  0.874    & 0.892    \\
		\hline
		\#. Parameters    & 95,107& 168,963& 21,123    &  94,979 \\
		
		\hline
		
		\hline
	\end{tabular}
	\label{table:ablation variants}
\end{table}

\begin{figure}[!htb]\footnotesize
	\setlength{\tabcolsep}{1pt}
	\centering
	\begin{tabular}{ccccccc}
		\!\!\!\!\!\! \includegraphics[width=.25\textwidth]{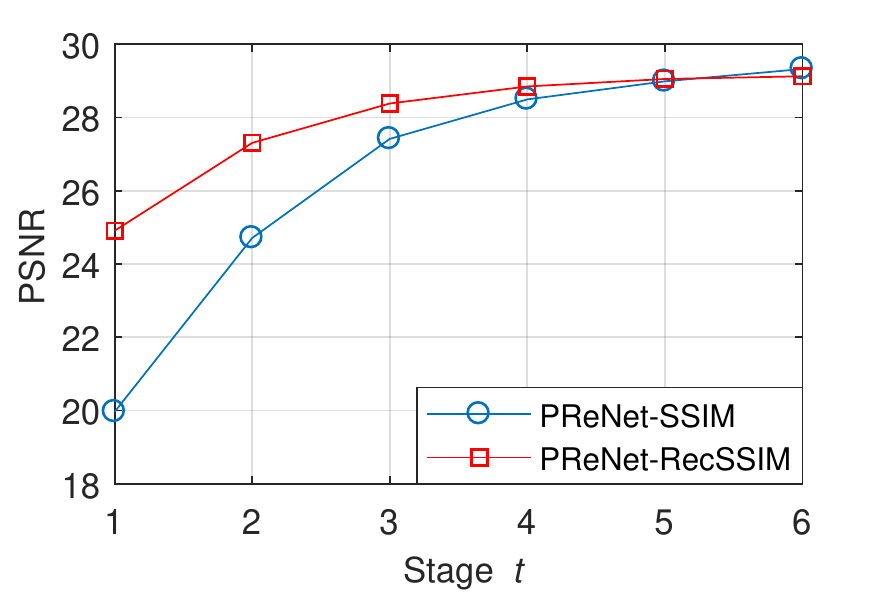} &
		\includegraphics[width=.25\textwidth]{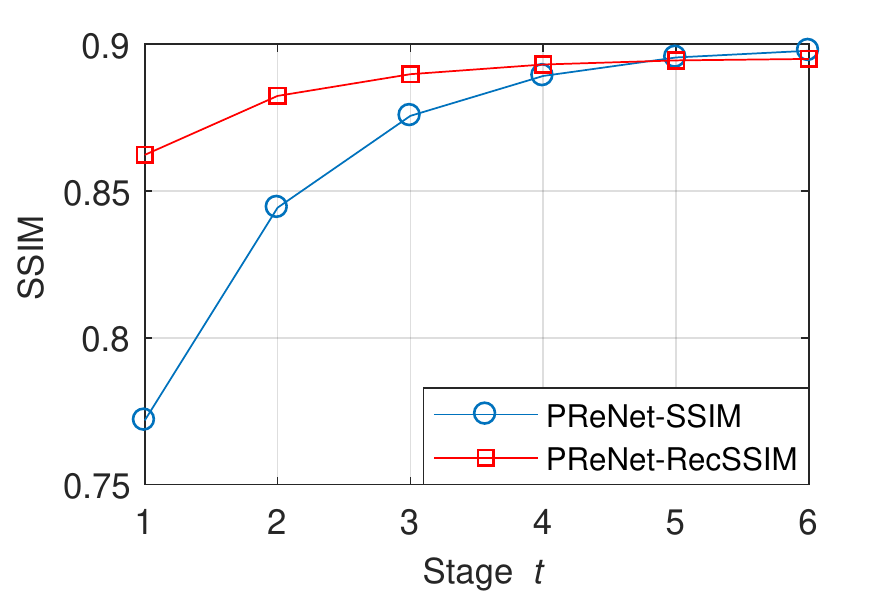}
	\end{tabular}
	\caption{Average PSNR and SSIM of PReNet-SSIM ($T=6$) and PReNet-RecSSIM ($T=6$) at stage $t=1,2,3,4,5,6$.}
	\label{fig:4states curve}
\end{figure}

\vspace{.05in}
\noindent \textbf{Single~v.s.~Recursive~Supervision.}
The negative SSIM loss can be imposed only on the final stage (PReNet-SSIM) in Eqn.~(\ref{eq:ssim loss}) or recursively on each stage (PReNet-RecSSIM) in Eqn.~(\ref{eq:multi loss}).
For PReNet-RecSSIM, we set $\lambda_t = 0.5\ (t = 1,2,...,5)$ and $\lambda_6 = 1.5$, where the tradeoff parameter for the final stage is larger than the others.
From Table~\ref{table:ablation loss}, PReNet-RecSSIM performs moderately inferior to PReNet-SSIM.
As shown in Fig.~\ref{fig:results loss}, the deraining results by PReNet-SSIM and PReNet-RecSSIM are visually indistinguishable.
The results indicate that a single loss on the final stage is sufficient to train progressive networks.
Furthermore, Fig.~\ref{fig:4states curve} shows the intermediate PSNR and SSIM results at each stage for PReNet-SSIM ($T$ = 6) and PReNet-RecSSIM ($T$ = 6).
It can be seen that PReNet-RecSSIM can achieve much better intermediate results than PReNet-SSIM, making PReNet-RecSSIM ($T$ = 6) very promising in computing resource constrained environments by stopping the inference at any stage $t$.


%
%
%

\subsubsection{Network Architecture}
In this subsection, we assess the effect of several key modules of progressive networks, including recurrent layer, multi-stage recursion, and intra-stage recursion.

\vspace{.05in}
\noindent \textbf{Recurrent Layer.} Using PReNet ($T = 6$), we test two types of recurrent layers, \ie, LSTM (PReNet-LSTM) and GRU (PReNet-GRU).
It can be seen from Table \ref{table:ablation more} that LSTM performs slightly better than GRU in terms of quantitative metrics, and thus is adopted as the default implementation of recurrent layer in our experiments.
We further compare progressive networks with and without recurrent layer, \ie, PReNet and PRN, in Table~\ref{table:ablation variants}, and obviously the introduction of recurrent layer does benefit the deraining performance in terms of PSNR and SSIM.

\begin{figure*}[!htb]\footnotesize
	\centering
	\setlength{\tabcolsep}{1pt}
	
	\begin{tabular}{cccccccccccccc}
		\includegraphics[width=.2\textwidth]{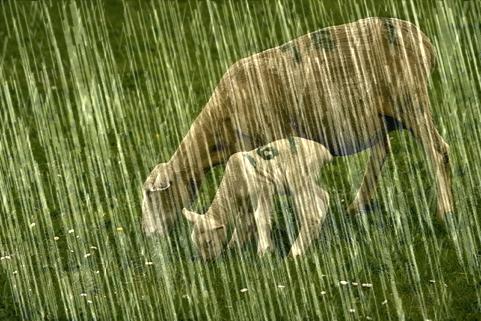}  &
		\includegraphics[width=.2\textwidth]{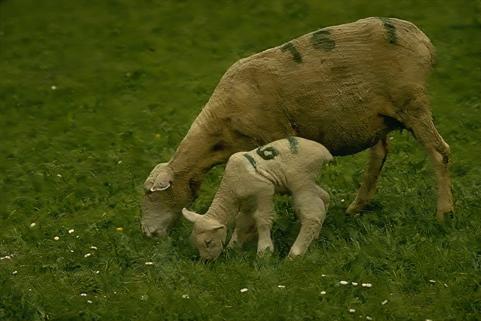} &
		\includegraphics[width=.2\textwidth]{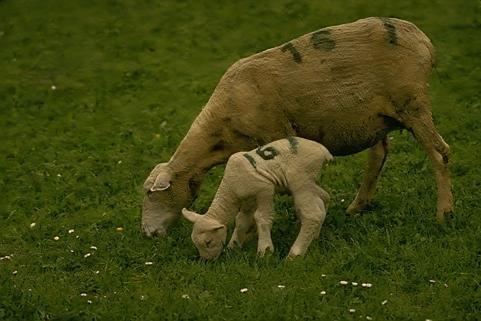}&
		\includegraphics[width=.2\textwidth]{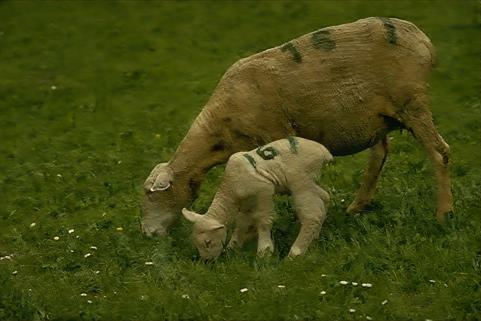}  &
		\includegraphics[width=.2\textwidth]{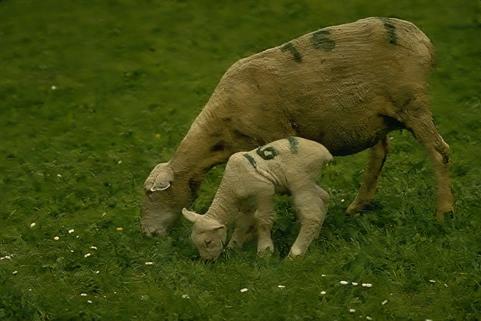} \\

		(a) Rainy image & (b) PRN & (c) PReNet & (d)PRN$_r$ & (e) PReNet$_r$ \\
		
	\end{tabular}
	\caption{Visual effects of recursive ResBlocks. The deraining results by PRN$_r$ and PReNet$_r$ are visually indistinguishable with those by PRN and PReNet.  }
	\label{fig:recursive results}
\end{figure*}

%
\vspace{.05in}
\noindent \textbf{Intra-stage Recursion.}
%
%
From Table~\ref{table:ablation variants}, intra-stage recursion, \ie, recursive ResBlocks, is introduced to significantly reduce the number of parameters of progressive networks, resulting in PRN$_r$ and PReNet$_r$.
As for deraining performance, it is reasonable to see that PRN and PReNet respectively achieve higher average PSNR and SSIM values than PRN$_r$ and PReNet$_r$.
But in terms of visual quality, PRN$_r$ and PReNet$_r$ are comparable with PRN and PReNet, as shown in Fig. \ref{fig:recursive results}.
%
%

\vspace{.05in}
\noindent \textbf{Recursive Stage Number $T$.}
Table \ref{table:ablation stages} lists the PSNR and SSIM values of four PReNet models with stages $T = 2,3,4,5,6,7$.
One can see that PReNet with more stages (from 2 stages to 6 stages) usually leads to higher average PSNR and SSIM values.
However, larger $T$ also makes PReNet more difficult to train.
When $T = 7$, PReNet$_7$ performs a little inferior to PReNet$_6$.
Thus, we set $T = 6$ in the following experiments.
%

\subsubsection{Effect of Network Input/Output}\label{sec:experiments input}

\vspace{.05in}
\noindent \textbf{Network Input.}
We also test a variant of PReNet by only taking $\mathbf{x}^{t-1}$ at each stage as input to $f_{in}$ (\ie, PReNet$_x$), where such strategy has been adopted in~\cite{yang2017deep,li2018recurrent}.
From Table~\ref{table:ablation more}, PReNet$_x$ is obviously inferior to PReNet in terms of both PSNR and SSIM, indicating the benefit of receiving $\mathbf{y}$ at each stage.


\vspace{.05in}
\noindent \textbf{Network Output.}
We consider two types of network outputs by incorporating residual learning formulation (\ie, PReNet in Table \ref{table:ablation more}) or not (\ie, PReNet-LSTM  in Table \ref{table:ablation more}).
From Table~\ref{table:ablation more}, residual learning can make a further contribution to performance gain.
It is worth noting that, benefited from progressive networks, it is feasible to
learn PReNet for directly predicting clean background from rainy image, and even PReNet-LSTM can achieve appealing deraining performance.


\subsection{Evaluation on Synthetic Datasets}\label{sec:synthetic}

\begin{table*}[!htb]\small
	\setlength{\tabcolsep}{2pt}
	\caption{Average PSNR and SSIM comparison on the synthetic datasets Rain100H \cite{yang2017deep}, Rain100L \cite{yang2017deep} and Rain12 \cite{li2016rain}. {\first{Red}}, \second{blue} and \third{cyan} colors are used to indicate top \first{$1^{\text{st}}$}, \second{$2^{\text{nd}}$} and \third{$3^{\text{rd}}$} rank, respectively.
		$^\triangleright$ means these metrics are copied from \cite{fan2018residual}.
		$^\circ$ means the metrics are directly computed based on the deraining images provided by the authors \cite{yang2017deep}.
		$^\star$ donates the method is re-trained with their default settings (\ie, all the 1800 training samples for Rain100H).
	}
	\centering
	\begin{tabular}{c|ccccc|cc|cc}
		\hline
		
		\hline
		Method  & GMM~\cite{li2016rain} & DDN~\cite{fu2017removing} & RGN~\cite{fan2018residual}$^\triangleright$  & JORDER~\cite{yang2017deep}$^\circ$ & RESCAN~\cite{li2018recurrent}$^\star$ & PRN  &PReNet & PRN$_r$  &PReNet$_r$  \\
		\hline
		Rain100H & 15.05/0.425  & 21.92/0.764  & 25.25/0.841  & 26.54/0.835  & \third{28.88}/0.866&   28.07/\third{0.884}   &\first{29.46}/\first{0.899}  &   27.43/0.874   &\second{28.98}/\second{0.892}\\
		Rain100L & 28.66/0.865  & 32.16/0.936  & 33.16/0.963  & 36.61/\third{0.974}  & ------     &   \third{36.99}/\second{0.977}   &\first{37.48}/\first{0.979}  &   36.11/0.973   & \second{37.10}/\second{0.977}\\
		
		Rain12   & 32.02/0.855  & 31.78/0.900  & 29.45/0.938  & 33.92/\third{0.953}  & ------     &   \third{36.62}/0.952   &\second{36.66}/\second{0.961} &  36.16/\second{0.961}   &\first{36.69}/\first{0.962}\\
		\hline
		
		\hline
	\end{tabular}
	\label{table:jorder dataset}
\end{table*}

\begin{figure*}[!htb]\footnotesize
	\centering
	\setlength{\tabcolsep}{0pt}
	
	\begin{tabular}{cclcclcclcclcclccl}

		\multicolumn{3}{c}{\includegraphics[width=.24\textwidth]{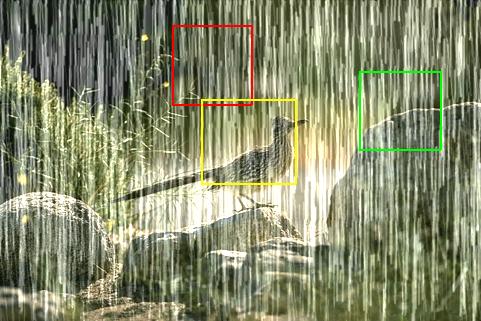}}\ &
		\multicolumn{3}{c}{\includegraphics[width=.24\textwidth]{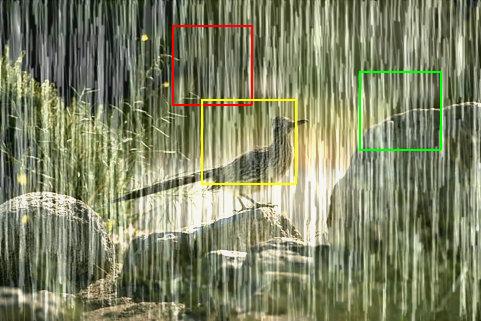}}\ &
		\multicolumn{3}{c}{\includegraphics[width=.24\textwidth]{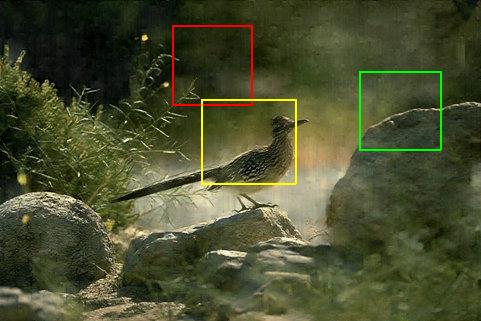}}\ &
		\multicolumn{3}{c}{\includegraphics[width=.24\textwidth]{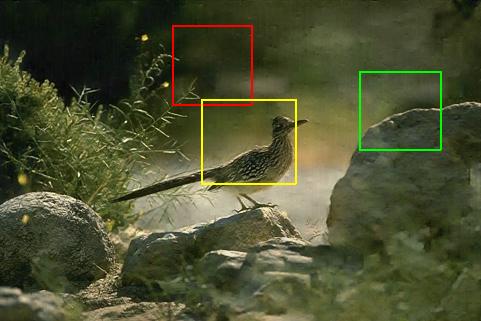}}\vspace{-2pt}\\
		\includegraphics[width=.08\textwidth]{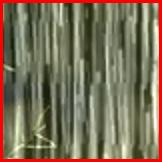} &
		\includegraphics[width=.08\textwidth]{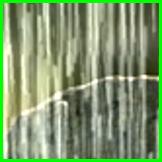} &
		\includegraphics[width=.08\textwidth]{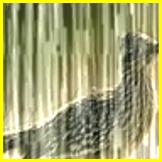}\ &
		\includegraphics[width=.08\textwidth]{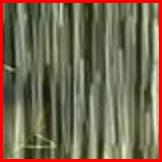}&
		\includegraphics[width=.08\textwidth]{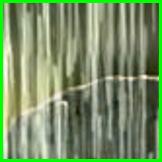}&
		\includegraphics[width=.08\textwidth]{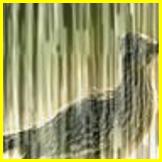}\ &
		\includegraphics[width=.08\textwidth]{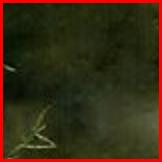}&
		\includegraphics[width=.08\textwidth]{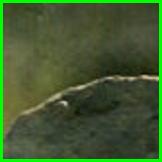}&
		\includegraphics[width=.08\textwidth]{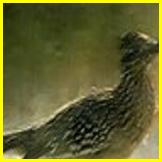}\ &
		\includegraphics[width=.08\textwidth]{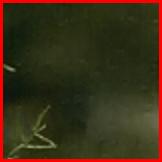}&
		\includegraphics[width=.08\textwidth]{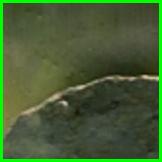}&
		\includegraphics[width=.08\textwidth]{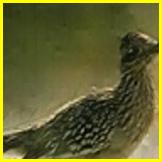}\\
		\multicolumn{3}{c}{Rainy image} &
		\multicolumn{3}{c}{GMM~\cite{li2016rain}} &
		\multicolumn{3}{c}{DDN~\cite{fu2017removing}} &
		\multicolumn{3}{c}{RESCAN~\cite{li2018recurrent}} \\
		\multicolumn{3}{c}{\includegraphics[width=.24\textwidth]{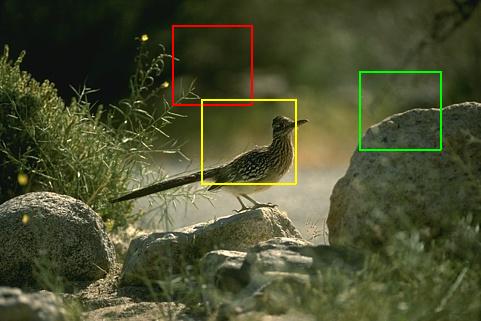}}\ &
		\multicolumn{3}{c}{\includegraphics[width=.24\textwidth]{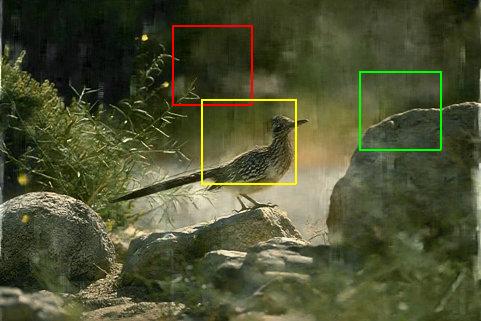}}\ &
		\multicolumn{3}{c}{\includegraphics[width=.24\textwidth]{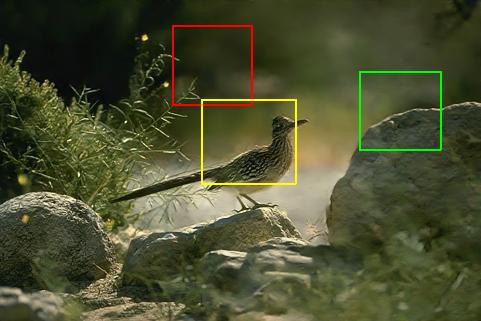}}\ &
		\multicolumn{3}{c}{\includegraphics[width=.24\textwidth]{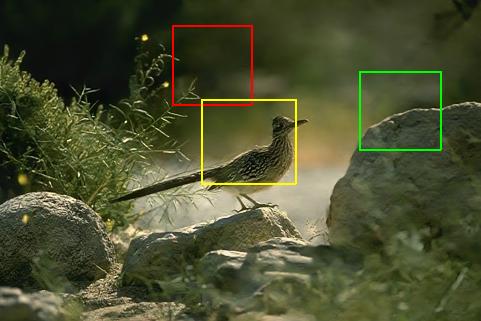}}\vspace{-2pt} \\
		\includegraphics[width=.08\textwidth]{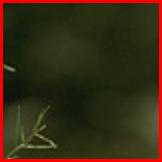} &
		\includegraphics[width=.08\textwidth]{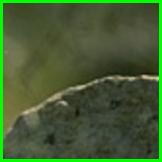} &
		\includegraphics[width=.08\textwidth]{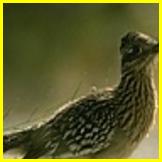}\  &
		\includegraphics[width=.08\textwidth]{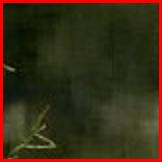} &
		\includegraphics[width=.08\textwidth]{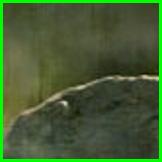} &
		\includegraphics[width=.08\textwidth]{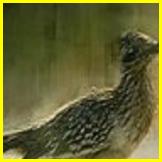}\ &
		\includegraphics[width=.08\textwidth]{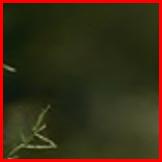} &
		\includegraphics[width=.08\textwidth]{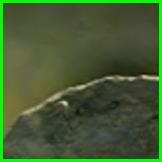} &
		\includegraphics[width=.08\textwidth]{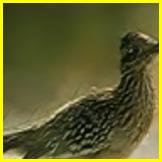}\ &
		\includegraphics[width=.08\textwidth]{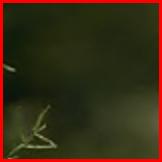} &
		\includegraphics[width=.08\textwidth]{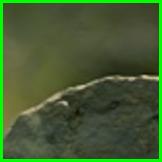} &
		\includegraphics[width=.08\textwidth]{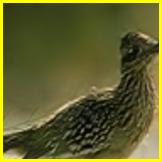} \\
		\multicolumn{3}{c}{Ground-truth} &
		\multicolumn{3}{c}{JORDER~\cite{yang2017deep}} &
		\multicolumn{3}{c}{PRN} &
		\multicolumn{3}{c}{PReNet} \\
		
	\end{tabular}
	\caption{Visual quality comparison on an image from Rain100H \cite{yang2017deep}. }
	\label{fig:results heavy and light}
\end{figure*}

\begin{figure*}[!htb]\footnotesize
	\centering
	\setlength{\tabcolsep}{0pt}
	
	\begin{tabular}{cclcclcclcclcclccl}
		\multicolumn{3}{c}{\includegraphics[width=.24\textwidth]{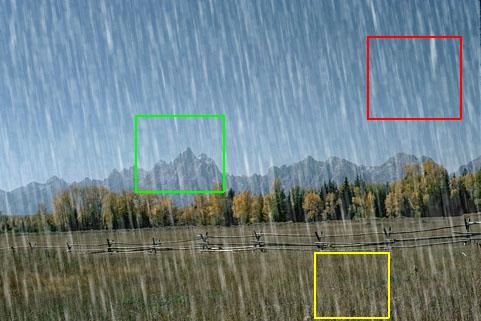}}\ &
		\multicolumn{3}{c}{\includegraphics[width=.24\textwidth]{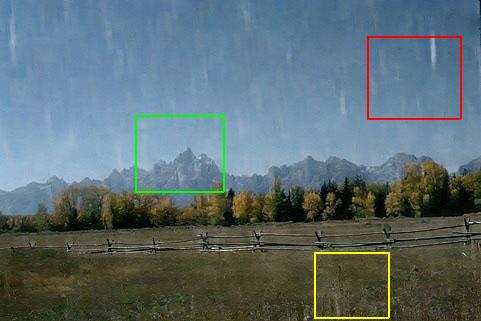}}\ &
		\multicolumn{3}{c}{\includegraphics[width=.24\textwidth]{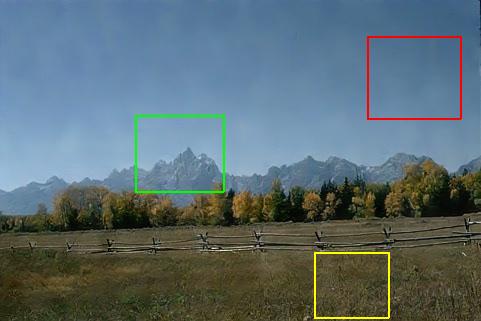}}\ &
		\multicolumn{3}{c}{\includegraphics[width=.24\textwidth]{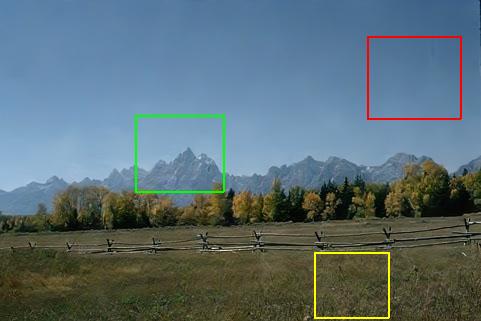}}\vspace{-2pt}\\
		\includegraphics[width=.08\textwidth]{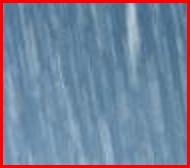} &
		\includegraphics[width=.08\textwidth]{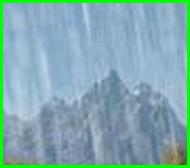} &
		\includegraphics[width=.08\textwidth]{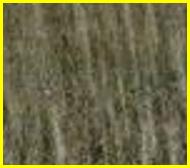}\ &
		\includegraphics[width=.08\textwidth]{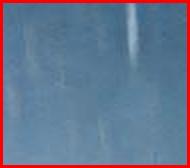}&
		\includegraphics[width=.08\textwidth]{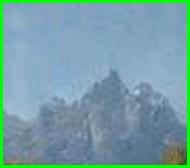}&
		\includegraphics[width=.08\textwidth]{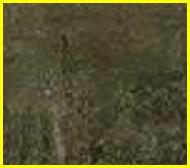}\ &
		\includegraphics[width=.08\textwidth]{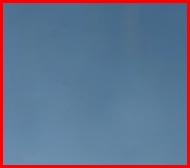}&
		\includegraphics[width=.08\textwidth]{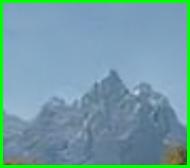}&
		\includegraphics[width=.08\textwidth]{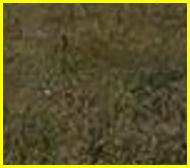}\ &
		\includegraphics[width=.08\textwidth]{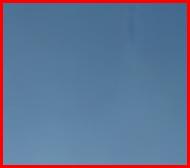}&
		\includegraphics[width=.08\textwidth]{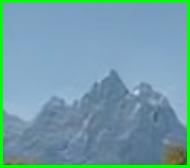}&
		\includegraphics[width=.08\textwidth]{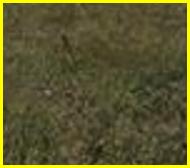}\\
		
		\multicolumn{3}{c}{Rainy image} &
		\multicolumn{3}{c}{DDN~\cite{fu2017removing}} &
		\multicolumn{3}{c}{PRN} &
		\multicolumn{3}{c}{PReNet} \\
	\end{tabular}
	\caption{Visual quality comparison on an image from Rain1400 \cite{fu2017removing}. }
	\label{fig:results r1400}
\end{figure*}

\begin{table}[!htb]\small
	\caption{Quantitative comparison on Rain1400 \cite{fu2017removing}.}
	\centering
	\setlength{\tabcolsep}{6pt}
	\begin{tabular}{c|c|cc|ccccc}
		\hline
		
		\hline
		Method & DDN~\cite{fu2017removing} &  PRN     & PReNet &  PRN$_r$   & PReNet$_r$  \\
		\hline
		PSNR   & 29.91   &  31.69   &  32.60   &  31.31   &  32.44   \\
		SSIM   & 0.910   &  0.941   &  0.946   &  0.937   &  0.944 \\
		\hline
		
		\hline
	\end{tabular}
	\label{table:fu dataset}
\end{table}

\begin{table*}[!htb]\small
	\setlength{\tabcolsep}{3pt}
	\caption{Comparison of running time (\emph{s})}
	
	\centering
	\begin{tabular}{c|ccc|ccccc}
		\hline
		
		\hline
		Image Size         & DDN~\cite{fu2017removing}  & JORDER~\cite{yang2017deep}   & RESCAN~\cite{li2018recurrent}  & PRN  & PReNet  \\
		\hline
		$500\times 500$   & 0.407     &  0.179       & 0.448   & 0.088  & 0.156  \\
		$1024\times 1024$ & 0.754     &  0.815       & 1.808   & 0.296  & 0.551  \\
		\hline
	
		\hline
	\end{tabular}
	\label{table:time}
\end{table*}

Our progressive networks are evaluated on three synthetic datasets, \ie, Rain100H \cite{yang2017deep}, Rain100L \cite{yang2017deep} and Rain12 \cite{li2016rain}.
Five competing methods are considered, including one traditional optimization-based method (GMM \cite{li2016rain}) and three state-of-the-art deep CNN-based models, \ie, DDN \cite{fu2017removing}, JORDER \cite{yang2017deep} and RESCAN \cite{li2018recurrent}, and one lightweight network (RGN \cite{fan2018residual}).
For heavy rainy images (Rain100H) and light rainy images (Rain100L), the models are respectively trained, and the models for light rain are directly applied on Rain12.
Since the source codes of RGN are not available, we adopt the numerical results reported in \cite{fan2018residual}.
As for JORDER, we directly compute average PSNR and SSIM on deraining results provided by the authors.
%
%
We re-train RESCAN \cite{li2018recurrent} for Rain100H with the default settings.
Besides, we have tried to train RESCAN for light rainy images, but the results are much inferior to the others.
So its results on Rain100L and Rain12 are not reported in our experiments.

Our PReNet achieves significant PSNR and SSIM gains over all the competing methods.
{\emph{We also note that for Rain100H, RESCAN \cite{li2018recurrent} is re-trained on the full 1,800 rainy images, the performance gain by our PReNet trained on the strict 1,254 rainy images is still notable. }}
Moreover, even PReNet$_r$ can perform better than all the competing methods.
From Fig.~\ref{fig:results heavy and light}, visible dark noises along rain directions can still be observed from the results by the other methods.
In comparison, the results by PRN and PReNet are visually more pleasing.

We further evaluate progressive networks on another dataset~\cite{fu2017removing} which includes 12,600 rainy images for training and 1,400 rainy images for testing (Rain1400).
From Table~\ref{table:fu dataset}, all the four versions of progressive networks outperform DDN in terms of PSNR and SSIM.
As shown in Fig.~\ref{fig:results r1400}, the visual quality improvement by our methods is also significant, while the result by DDN still contains visible rain streaks.

Table \ref{table:time} lists the running time of different methods based on a computer equipped with a NVIDIA GTX 1080Ti GPU.
We only give the running time of DDN \cite{fu2017removing}, JORDER \cite{yang2017deep} and RESCAN \cite{li2018recurrent}, due to the codes of the other competing methods are not available.
We note that the running time of DDN \cite{fu2017removing} takes the separation of details layer into account.
Unsurprisingly, PRN and PReNet are much more efficient due to its simple network architecture.

\subsection{Evaluation on Real Rainy Images}

\begin{figure*}[!htb]\footnotesize
	\centering
	\setlength{\tabcolsep}{0pt}
	
	\begin{tabular}{cclcclcclcclcclccl}
		\multicolumn{3}{c}{\includegraphics[width=.24\textwidth]{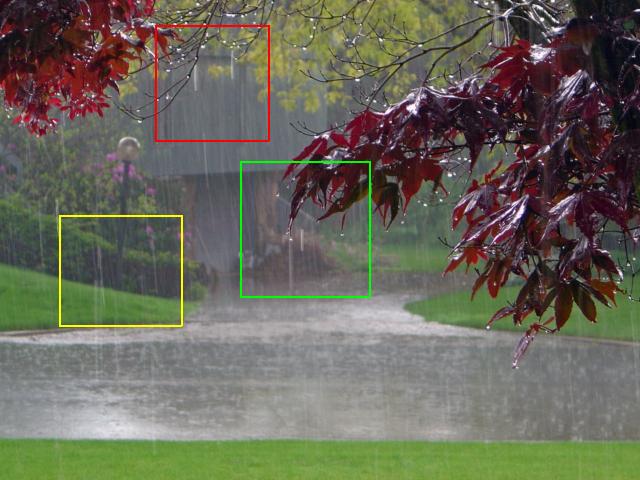}}\ &
		\multicolumn{3}{c}{\includegraphics[width=.24\textwidth]{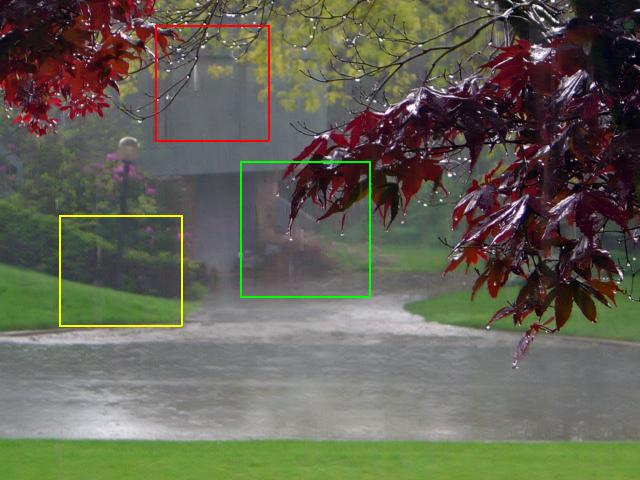}}\ &
		\multicolumn{3}{c}{\includegraphics[width=.24\textwidth]{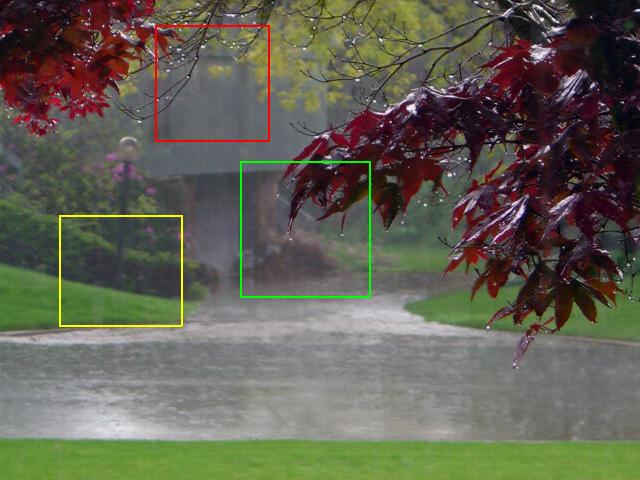}}\ &
		\multicolumn{3}{c}{\includegraphics[width=.24\textwidth]{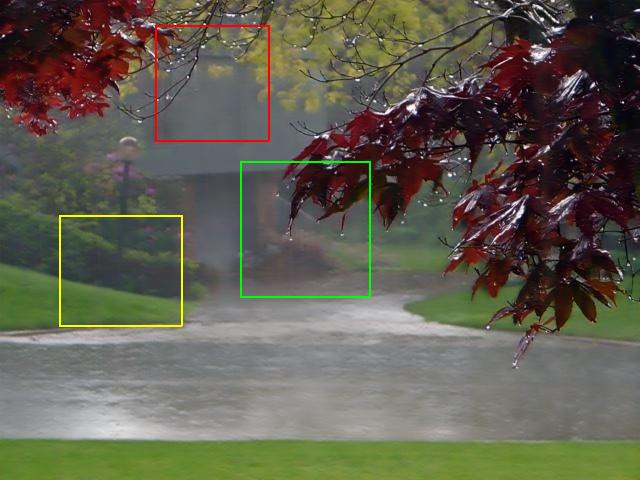}}\vspace{-2pt}\\
		\includegraphics[width=.08\textwidth]{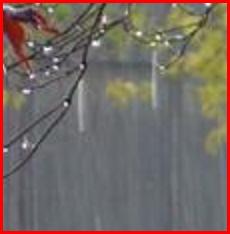} &
		\includegraphics[width=.08\textwidth]{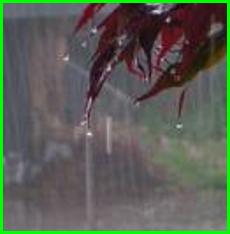} &
		\includegraphics[width=.08\textwidth]{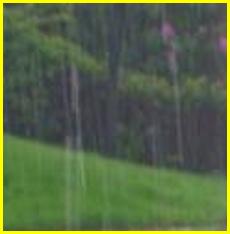}\ &
		\includegraphics[width=.08\textwidth]{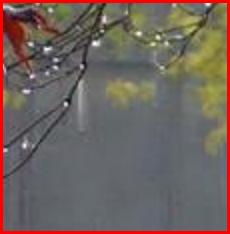}&
		\includegraphics[width=.08\textwidth]{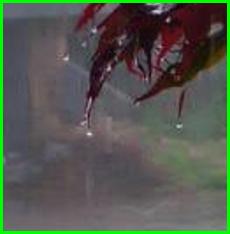}&
		\includegraphics[width=.08\textwidth]{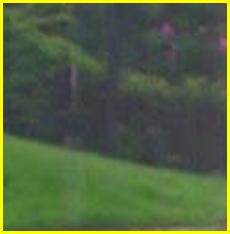}\ &
		\includegraphics[width=.08\textwidth]{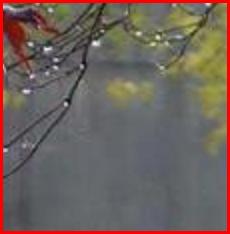}&
		\includegraphics[width=.08\textwidth]{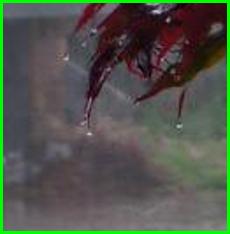}&
		\includegraphics[width=.08\textwidth]{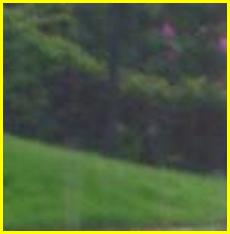}\ &
		\includegraphics[width=.08\textwidth]{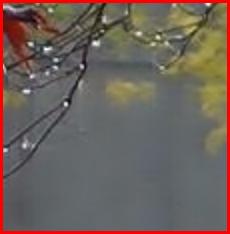}&
		\includegraphics[width=.08\textwidth]{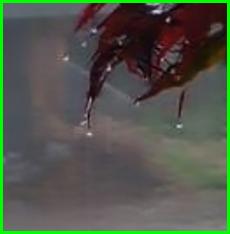}&
		\includegraphics[width=.08\textwidth]{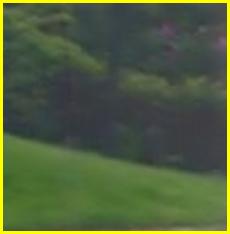}\\
		
		\multicolumn{3}{c}{\includegraphics[width=.24\textwidth]{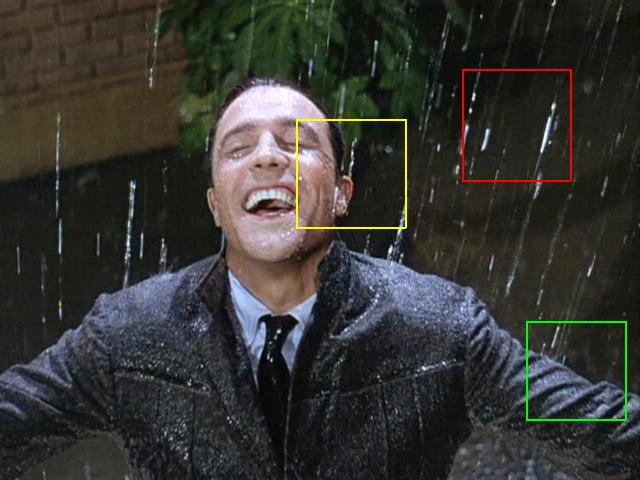}}\ &
		\multicolumn{3}{c}{\includegraphics[width=.24\textwidth]{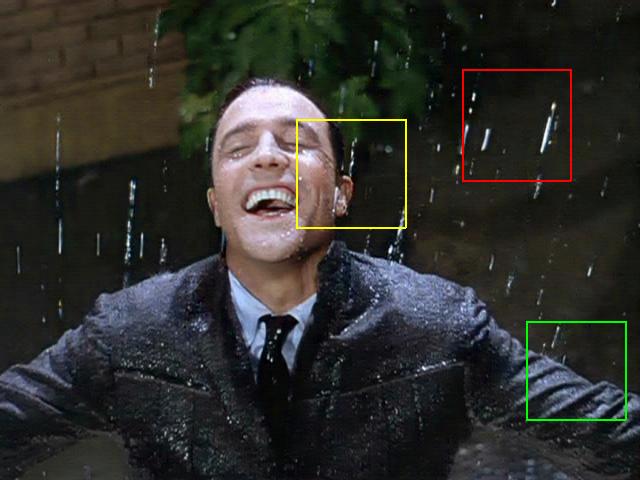}}\ &
		\multicolumn{3}{c}{\includegraphics[width=.24\textwidth]{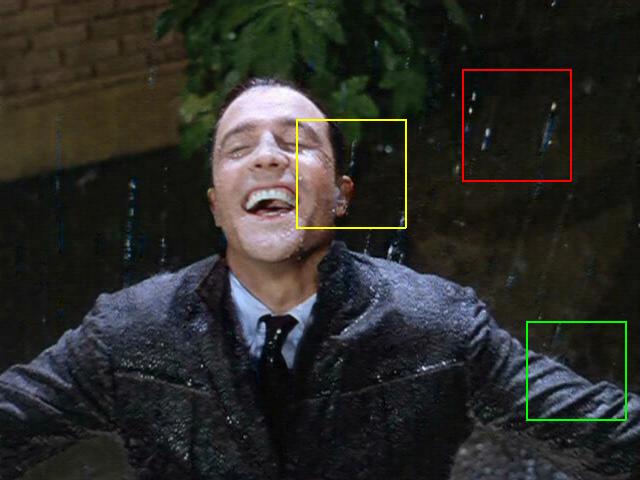}}\ &
		\multicolumn{3}{c}{\includegraphics[width=.24\textwidth]{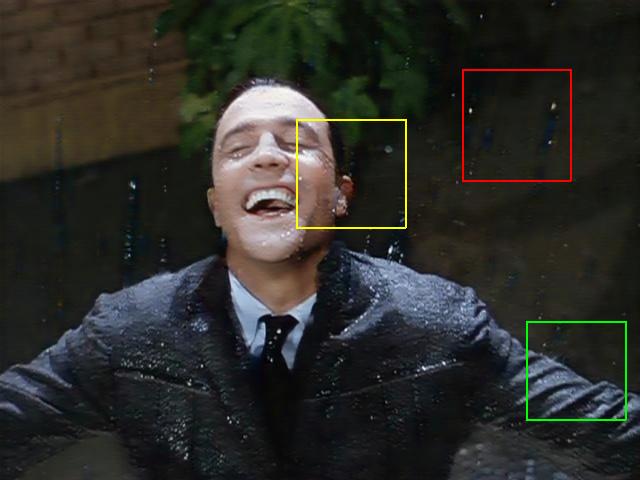}}\vspace{-2pt}\\
		\includegraphics[width=.08\textwidth]{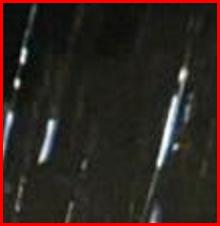} &
		\includegraphics[width=.08\textwidth]{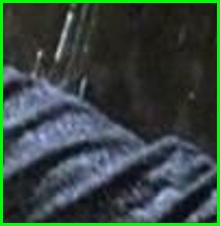} &
		\includegraphics[width=.08\textwidth]{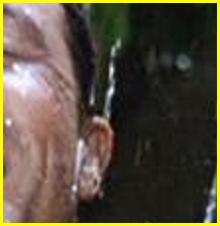}\ &
		\includegraphics[width=.08\textwidth]{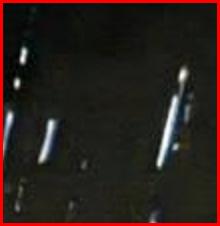}&
		\includegraphics[width=.08\textwidth]{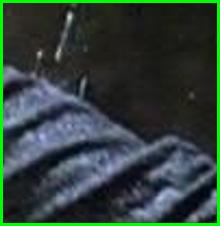}&
		\includegraphics[width=.08\textwidth]{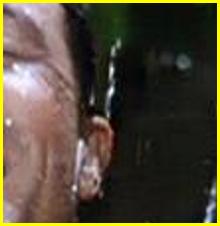}\ &
		\includegraphics[width=.08\textwidth]{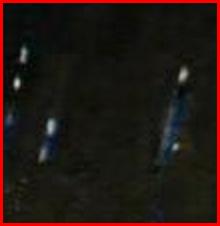}&
		\includegraphics[width=.08\textwidth]{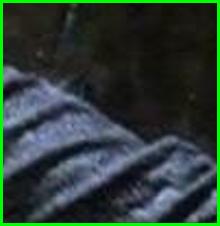}&
		\includegraphics[width=.08\textwidth]{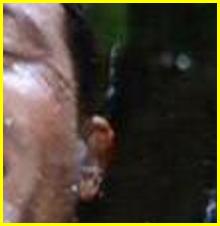}\ &
		\includegraphics[width=.08\textwidth]{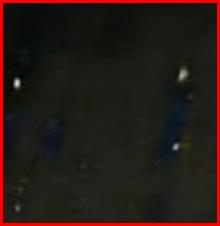}&
		\includegraphics[width=.08\textwidth]{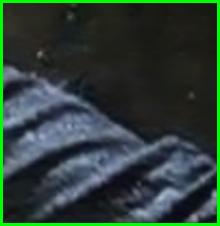}&
		\includegraphics[width=.08\textwidth]{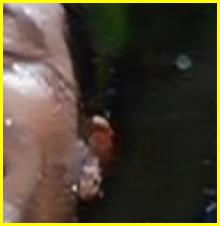}\\
		\multicolumn{3}{c}{Rainy image} &
		\multicolumn{3}{c}{DDN~\cite{fu2017removing}} &
		\multicolumn{3}{c}{JORDER~\cite{yang2017deep}} &
		\multicolumn{3}{c}{PReNet} \\
	\end{tabular}
	\caption{Visual quality comparison on two real rainy images.  }
	\label{fig:results real}
\end{figure*}

Using two real rainy images in Fig.~\ref{fig:results real}, we compare PReNet with two state-of-the-art deep methods, \ie, JORDER \cite{yang2017deep} and DDN \cite{fu2017removing}.
It can be seen that PReNet and JORDER perform better than DDN in removing rain streaks.
For the first image, rain streaks remain visible in the result by DDN, while PReNet and JORDER can generate satisfying deraining results.
For the second image, there are more or less rain streaks in the results by DDN and JORDER, while the result by PReNet is more clear.

\subsection{Evaluation on Real Rainy Videos}

\begin{figure*}[!htb]\footnotesize
	\centering
	\setlength{\tabcolsep}{0pt}
	
	\begin{tabular}{cclcclcclcclcclccl}

		\multicolumn{3}{c}{\includegraphics[width=.24\textwidth]{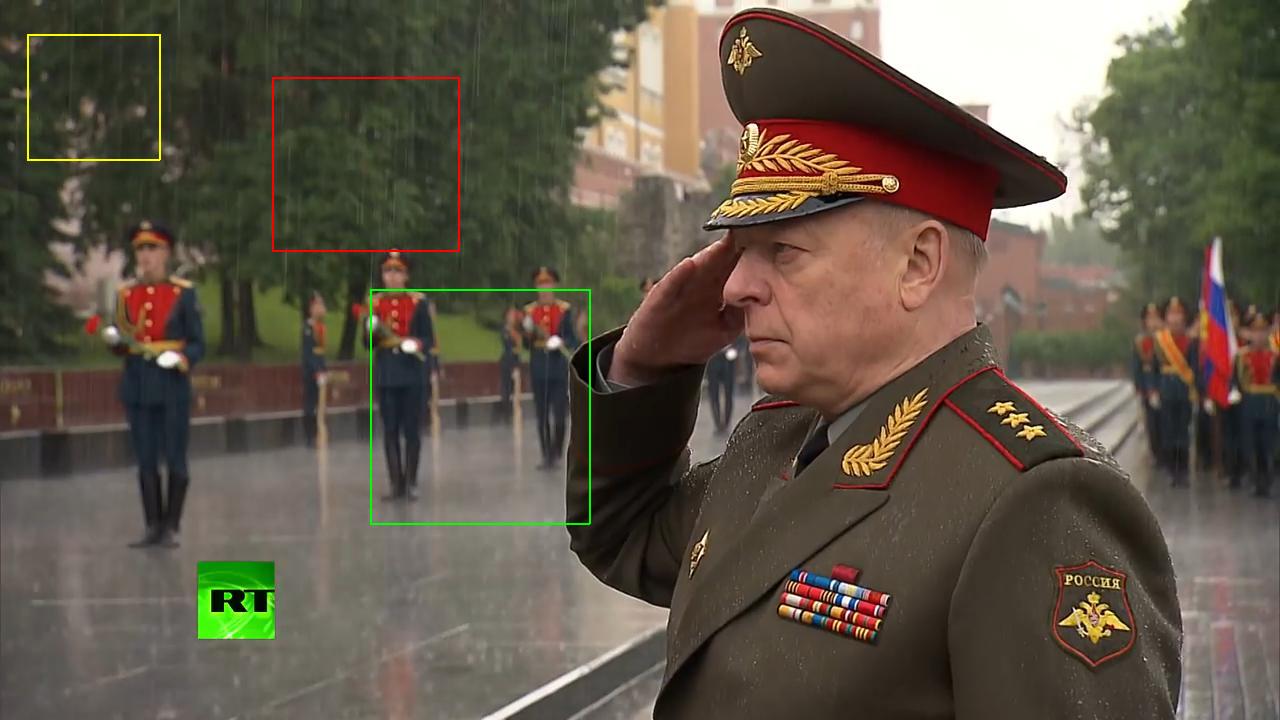}}\vspace{-2pt}\ &
		\multicolumn{3}{c}{\includegraphics[width=.24\textwidth]{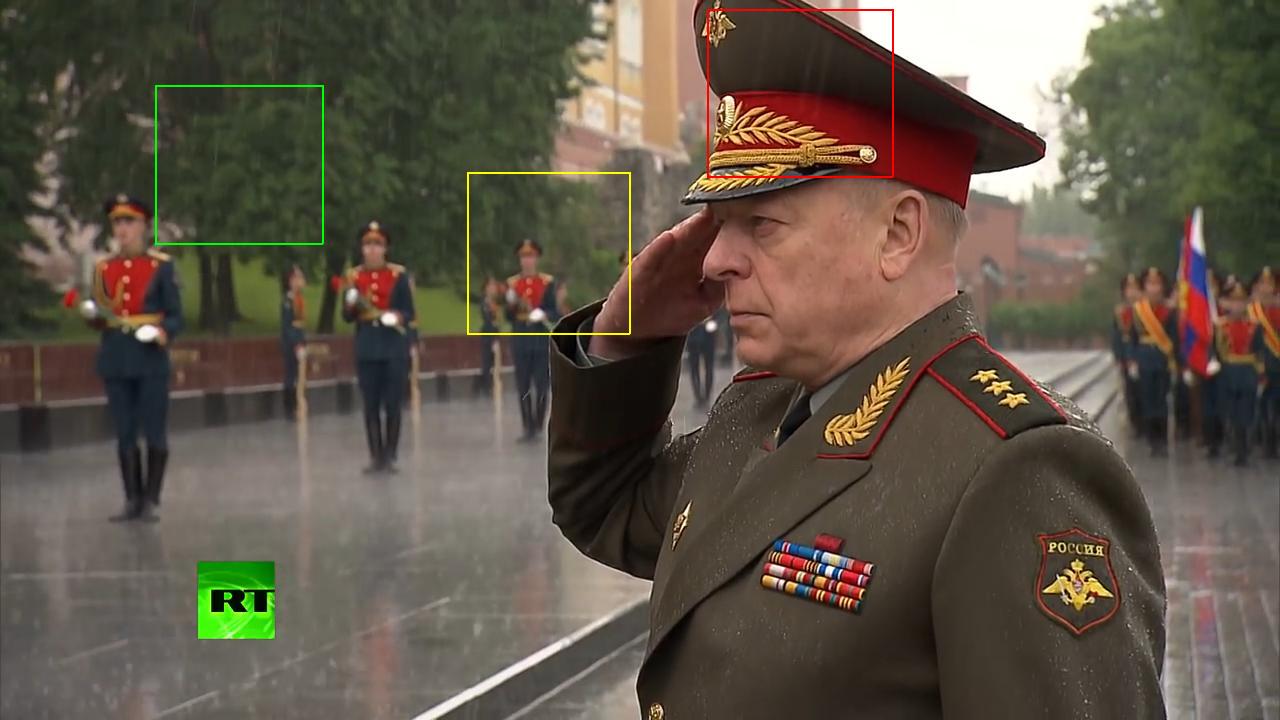}}\ &
		\multicolumn{3}{c}{\includegraphics[width=.24\textwidth]{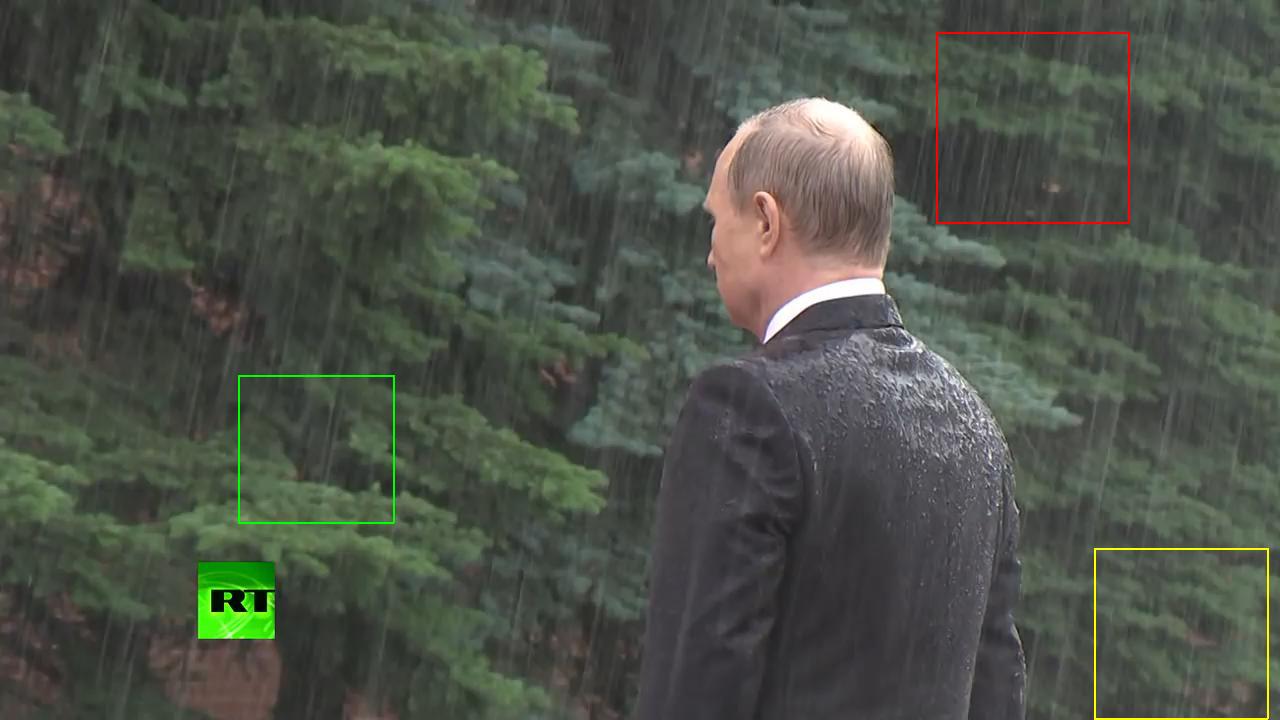}}\ &
		\multicolumn{3}{c}{\includegraphics[width=.24\textwidth]{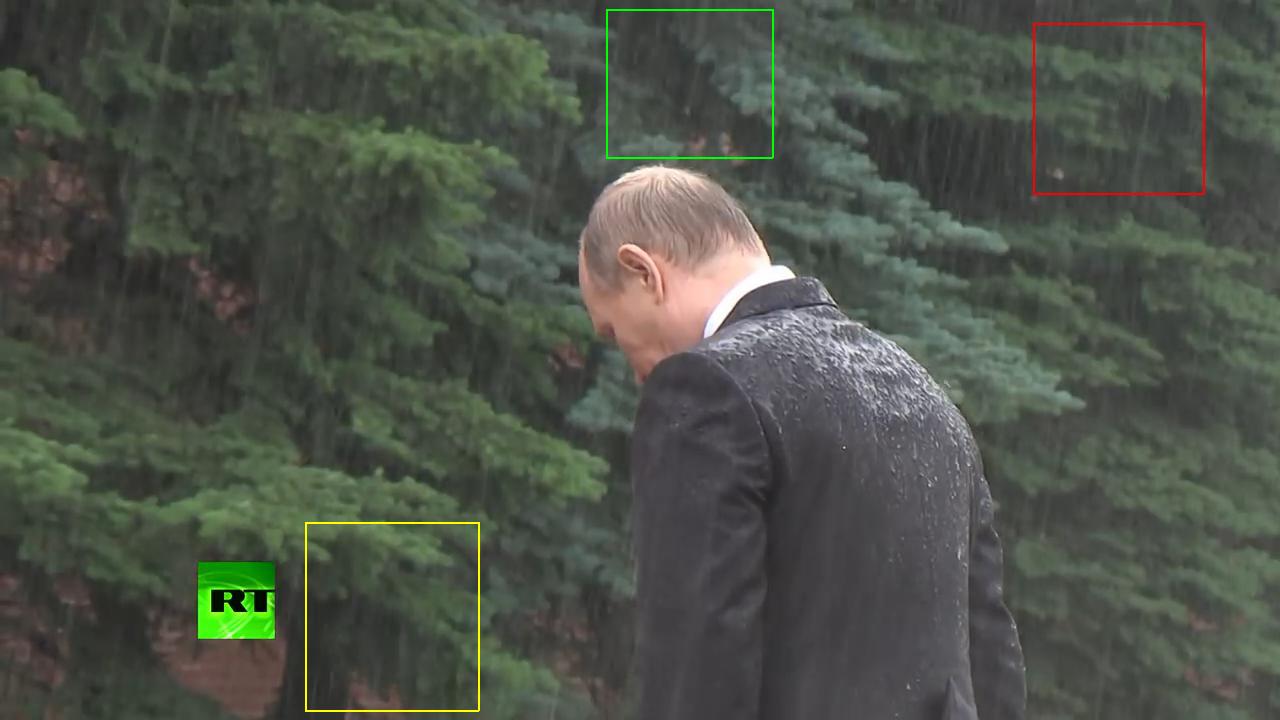}}\\
		\includegraphics[width=.08\textwidth]{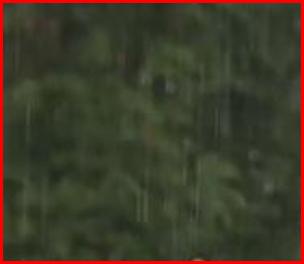} &
		\includegraphics[width=.08\textwidth]{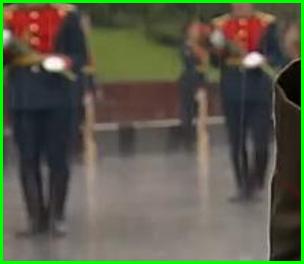} &
		\includegraphics[width=.08\textwidth]{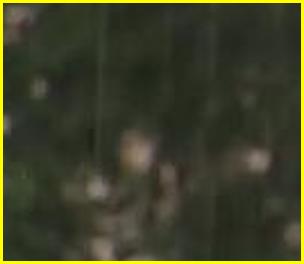}\ &
		\includegraphics[width=.08\textwidth]{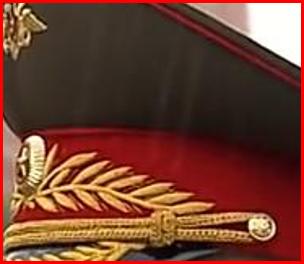}&
		\includegraphics[width=.08\textwidth]{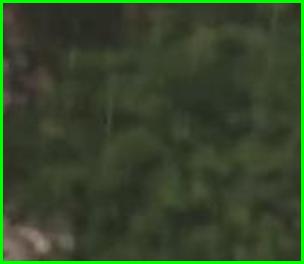}&
		\includegraphics[width=.08\textwidth]{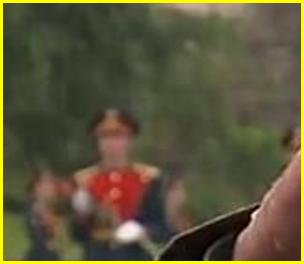}\ &
		\includegraphics[width=.08\textwidth]{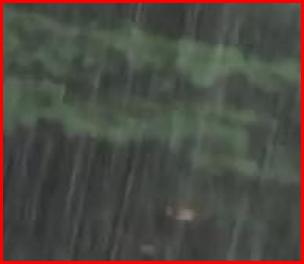}&
		\includegraphics[width=.08\textwidth]{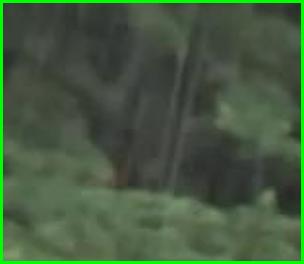}&
		\includegraphics[width=.08\textwidth]{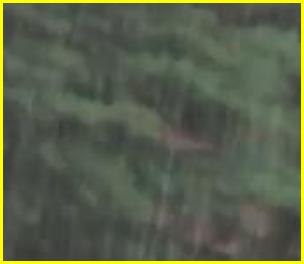}\ &
		\includegraphics[width=.08\textwidth]{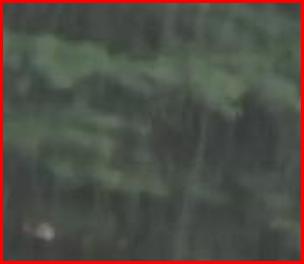}&
		\includegraphics[width=.08\textwidth]{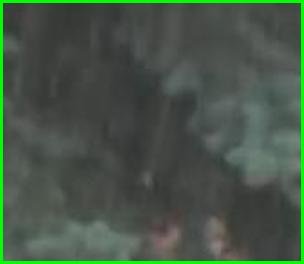}&
		\includegraphics[width=.08\textwidth]{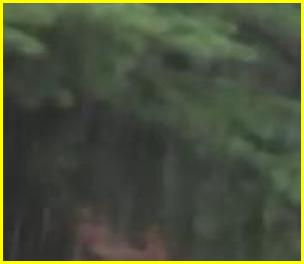}\\
		\multicolumn{3}{c}{\includegraphics[width=.24\textwidth]{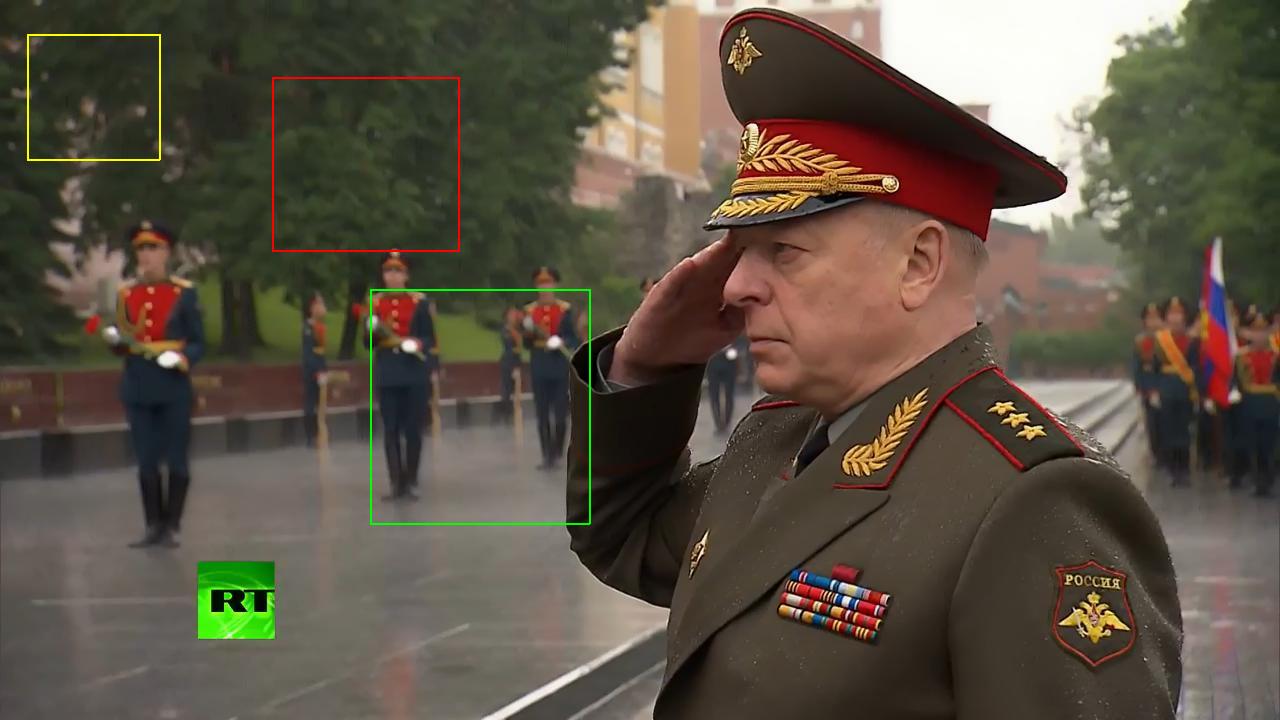}}\ &
		\multicolumn{3}{c}{\includegraphics[width=.24\textwidth]{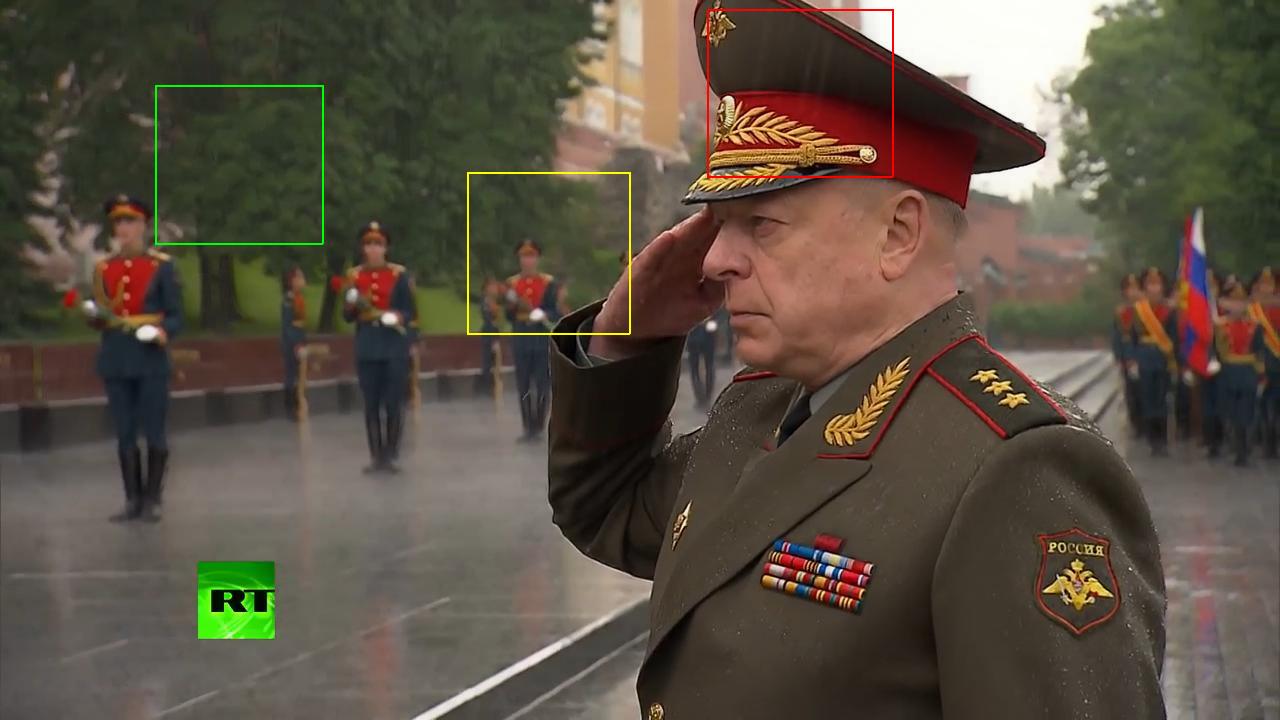}}\ &
		\multicolumn{3}{c}{\includegraphics[width=.24\textwidth]{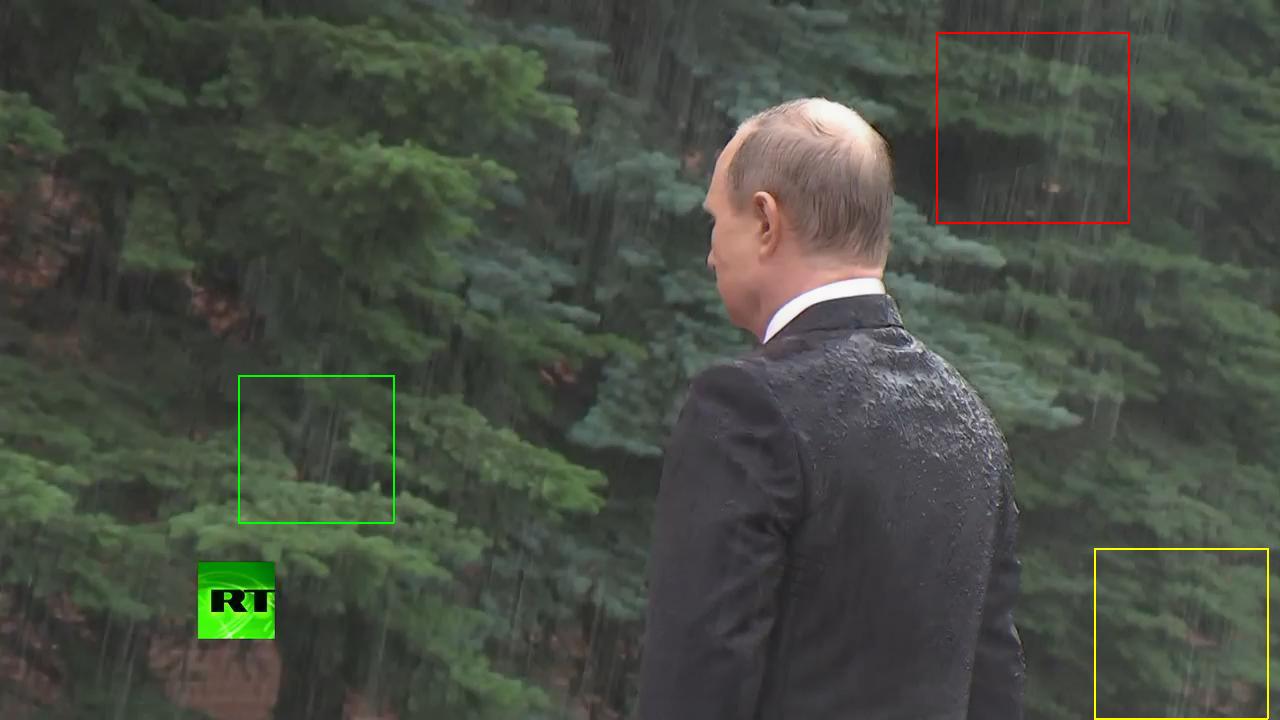}}\ &
		\multicolumn{3}{c}{\includegraphics[width=.24\textwidth]{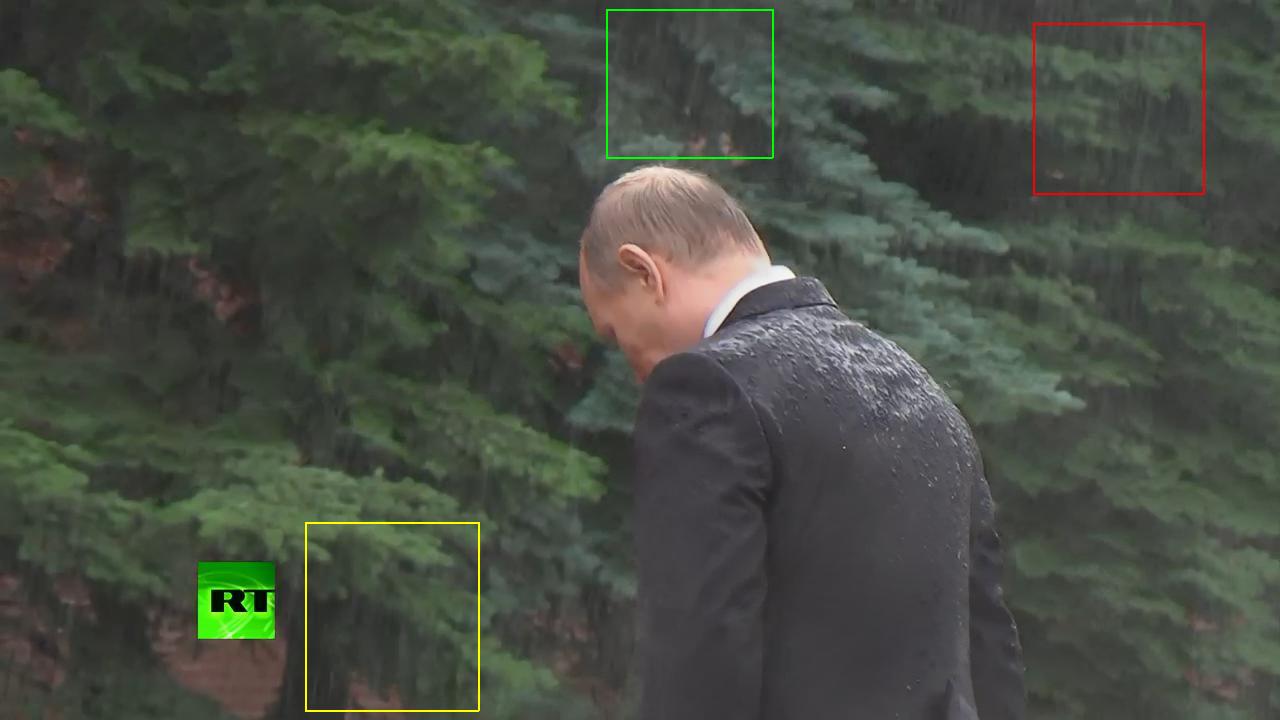}}\vspace{-2pt}\\
		\includegraphics[width=.08\textwidth]{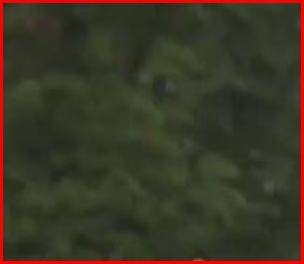} &
		\includegraphics[width=.08\textwidth]{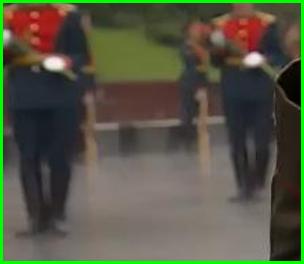} &
		\includegraphics[width=.08\textwidth]{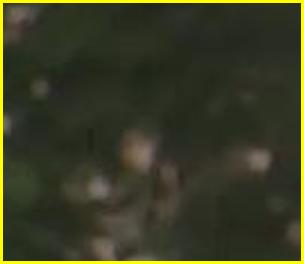}\ &
		\includegraphics[width=.08\textwidth]{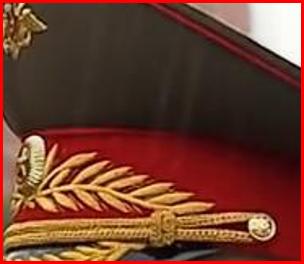}&
		\includegraphics[width=.08\textwidth]{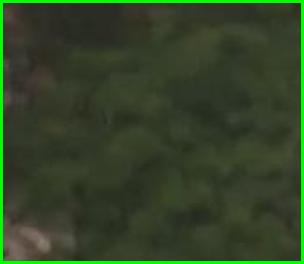}&
		\includegraphics[width=.08\textwidth]{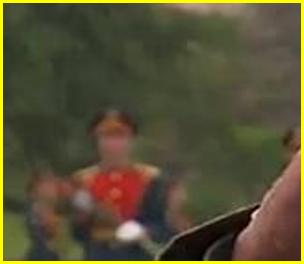}\ &
		\includegraphics[width=.08\textwidth]{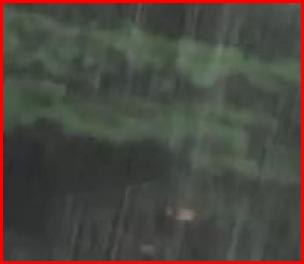}&
		\includegraphics[width=.08\textwidth]{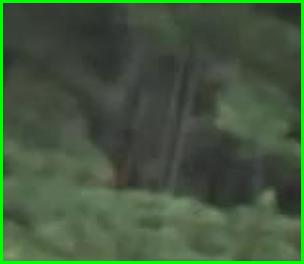}&
		\includegraphics[width=.08\textwidth]{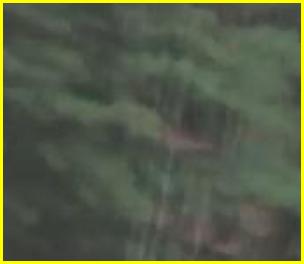}\ &
		\includegraphics[width=.08\textwidth]{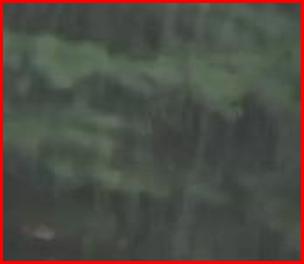}&
		\includegraphics[width=.08\textwidth]{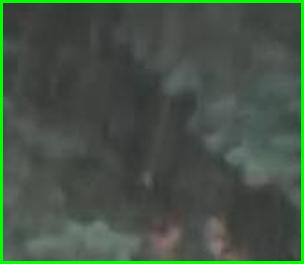}&
		\includegraphics[width=.08\textwidth]{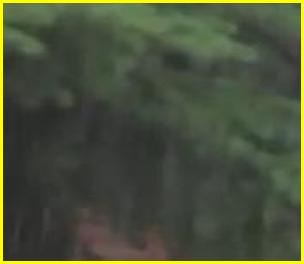}\\
		\multicolumn{3}{c}{\includegraphics[width=.24\textwidth]{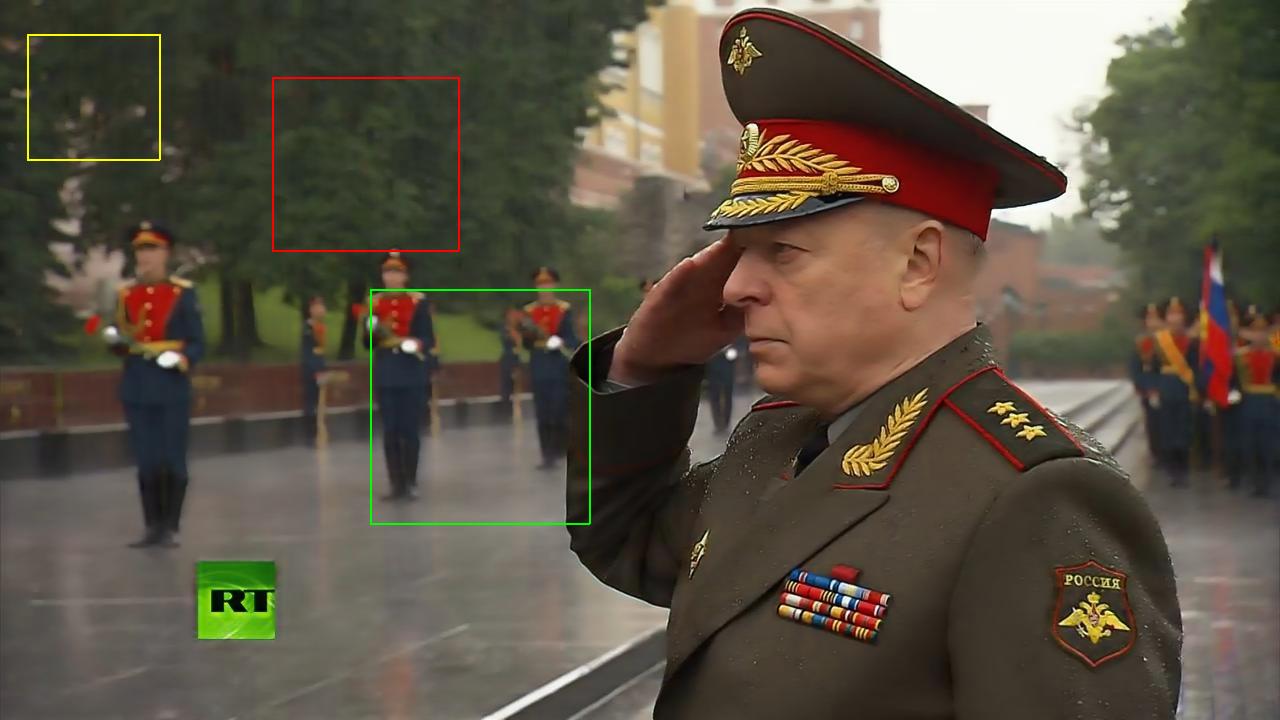}}\ &
		\multicolumn{3}{c}{\includegraphics[width=.24\textwidth]{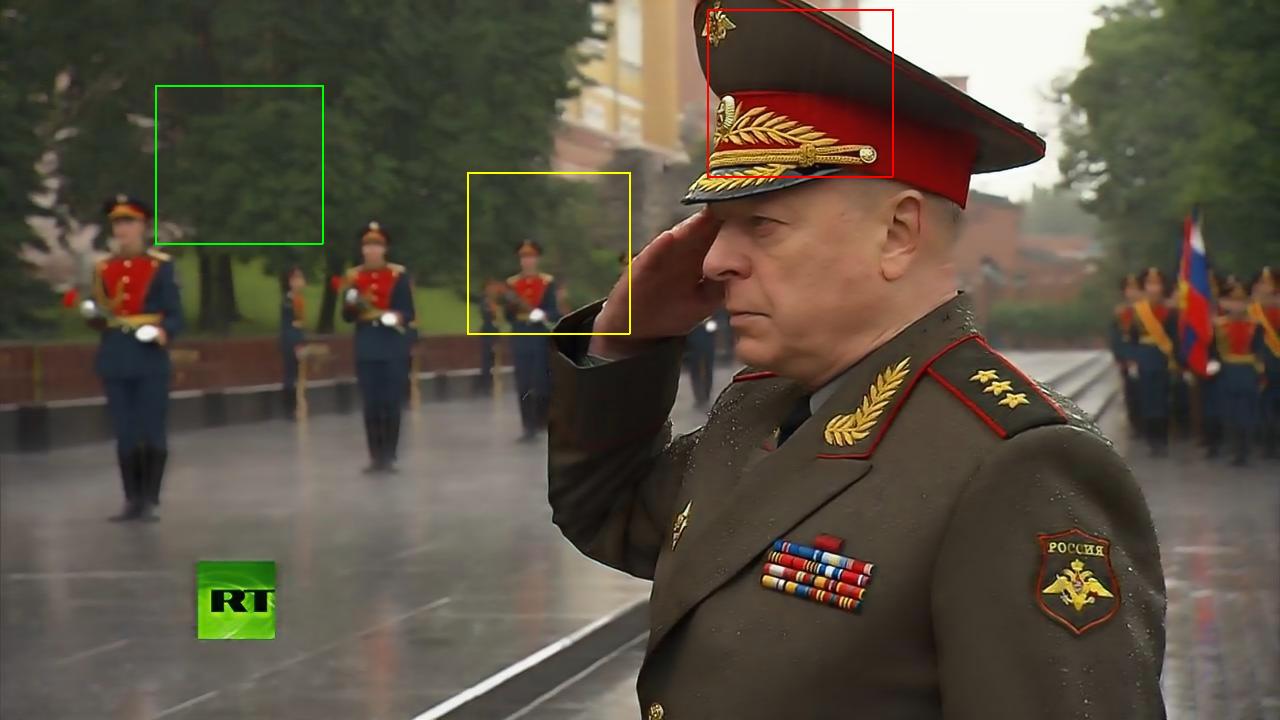}}\ &
		\multicolumn{3}{c}{\includegraphics[width=.24\textwidth]{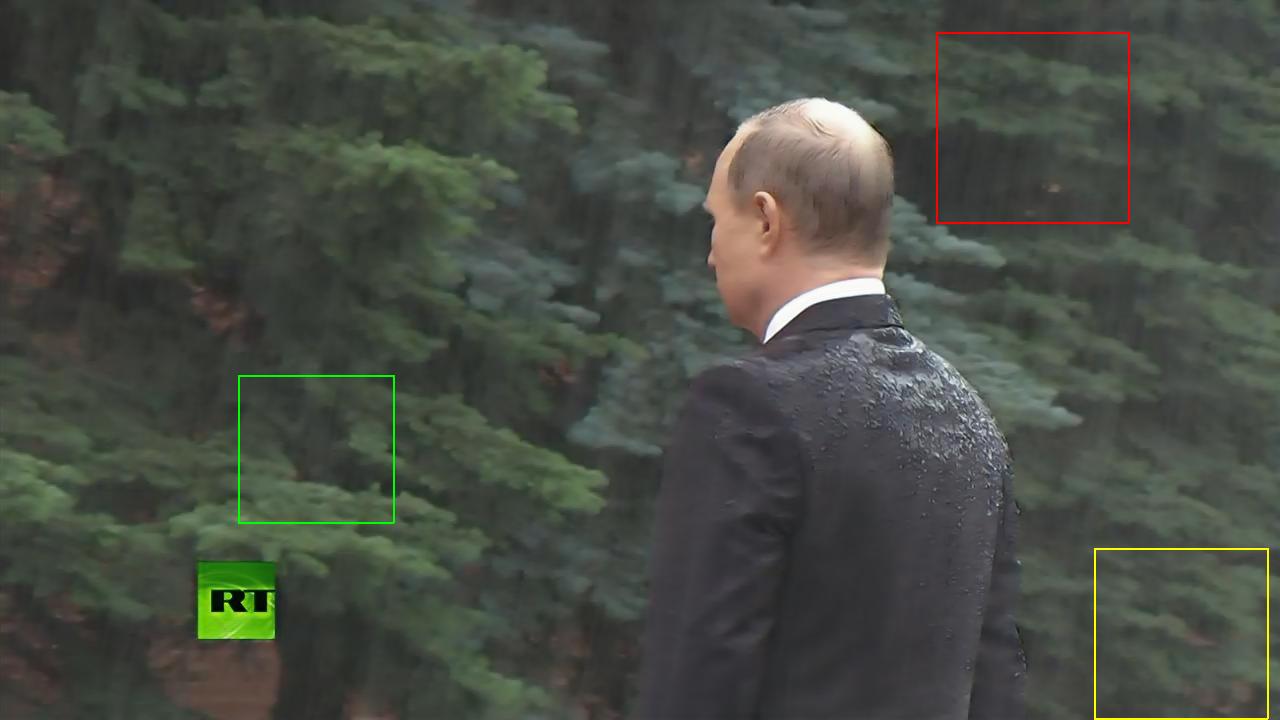}}\ &
		\multicolumn{3}{c}{\includegraphics[width=.24\textwidth]{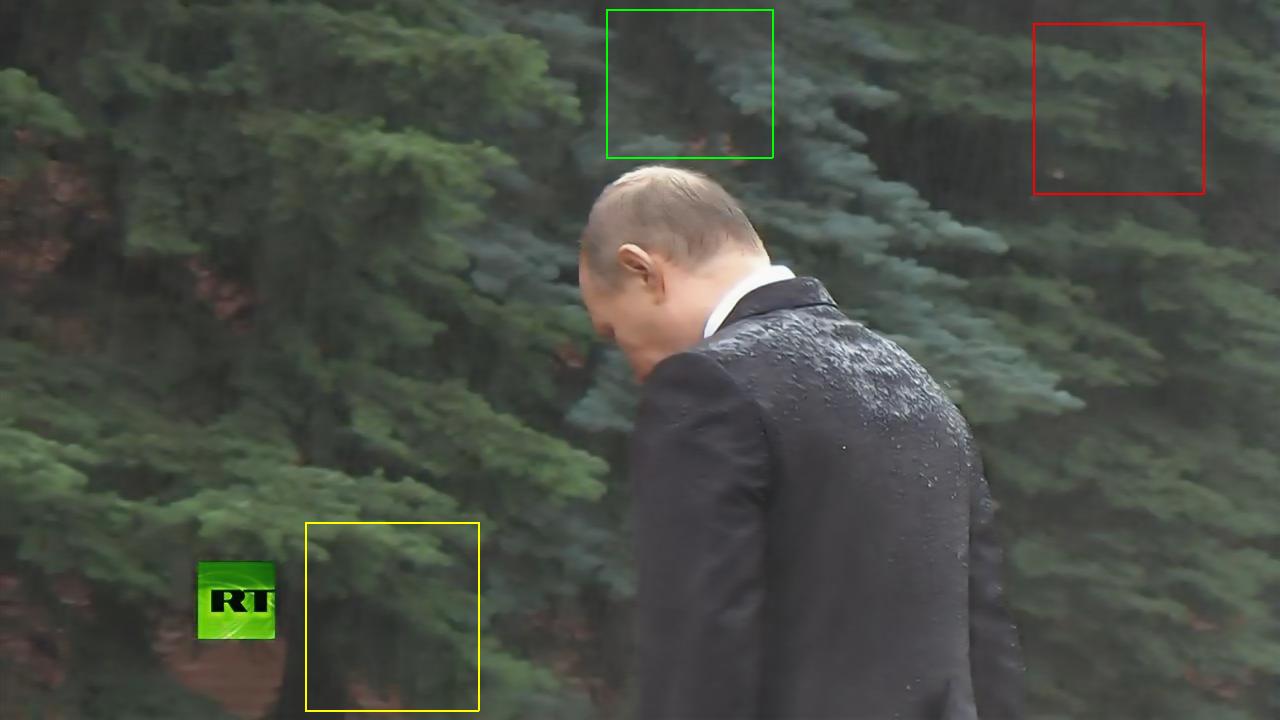}}\vspace{-2pt}\\
		\includegraphics[width=.08\textwidth]{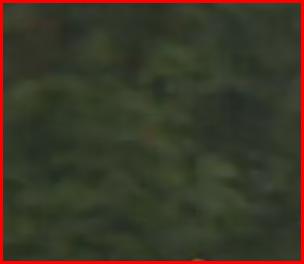} &
		\includegraphics[width=.08\textwidth]{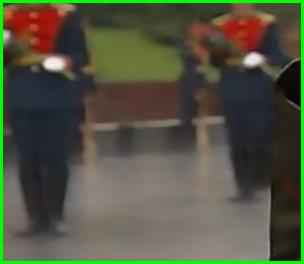} &
		\includegraphics[width=.08\textwidth]{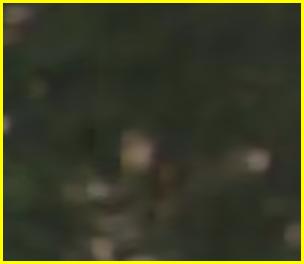}\ &
		\includegraphics[width=.08\textwidth]{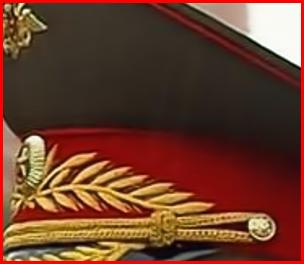}&
		\includegraphics[width=.08\textwidth]{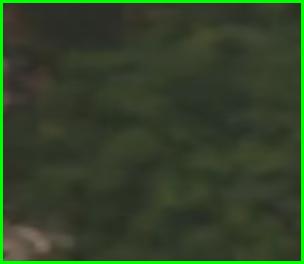}&
		\includegraphics[width=.08\textwidth]{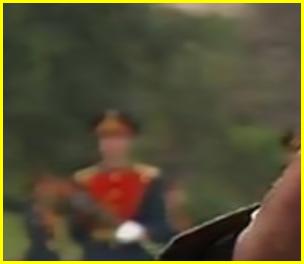}\ &
		\includegraphics[width=.08\textwidth]{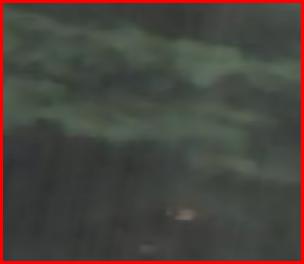}&
		\includegraphics[width=.08\textwidth]{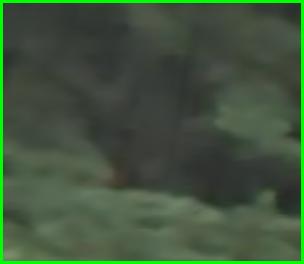}&
		\includegraphics[width=.08\textwidth]{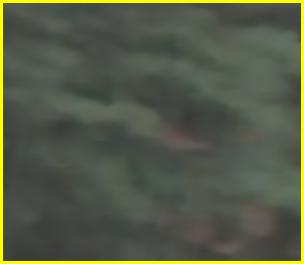}\ &
		\includegraphics[width=.08\textwidth]{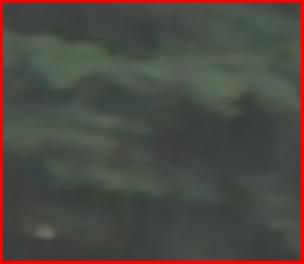}&
		\includegraphics[width=.08\textwidth]{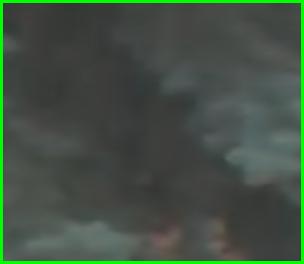}&
		\includegraphics[width=.08\textwidth]{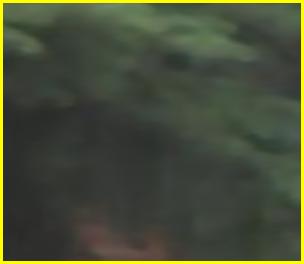}\\
		\multicolumn{3}{c}{Frame \#510} &
		\multicolumn{3}{c}{Frame \#571} &
		\multicolumn{3}{c}{Frame \#572} &
		\multicolumn{3}{c}{Frame \#640} \\
	\end{tabular}
	\caption{Visual quality comparison on a real rainy video. The first row is rainy frames, the second row is the results by FastDerain~\cite{jiang2017novel} and the third row is the results by PReNet. }
	\label{fig:results real video}
\end{figure*}

Finally, PReNet is adopted to process a rainy video in a frame-by-frame manner, and is compared with state-of-the-art video deraining method, \ie, FastDerain~\cite{jiang2017novel}.
As shown in Fig. \ref{fig:results real video}, for frame \#510, both FastDerain and our PReNet can remove all the rain streaks, indicating the performance of PReNet even without the help of temporal consistency.
However, FastDerain fails in switching frames, since it is developed by exploiting the consistency of adjacent frames.
As a result, for frame \#571, \#572 and 640, rain streaks are remained in the results by FastDerain, while our PReNet performs favorably and is not affected by switching frames and accumulation error.
%

\section{Conclusion}

In this paper, a better and simpler baseline network is presented for single image deraining.
Instead of deeper and complex networks, we find that the simple combination of ResNet and multi-stage recursion, \ie, PRN, can result in favorable performance.
Moreover, the deraining performance can be further boosted by the inclusion of recurrent layer, and stage-wise result is also taken as input to each ResNet, resulting in our PReNet model.
Furthermore, the network parameters can be reduced by incorporating inter- and intra-stage recursive computation (PRN$_r$ and PReNet$_r$).
And our progressive deraining networks can be readily trained with single negative SSIM or MSE loss.
Extensive experiments validate the superiority of our PReNet and PReNet$_r$ on synthetic and real rainy images in comparison to state-of-the-art deraining methods.
Taking their simplicity, effectiveness and efficiency into account, it is also appealing to
exploit our models as baselines when developing new deraining networks.


{\small
\bibliographystyle{ieee}
\bibliography{reference}
}

\end{document}